%% file: main.tex
\documentclass{article} 
\usepackage[final]{colm2026_conference}

\usepackage{microtype}
\usepackage{hyperref}
\usepackage{url}

\usepackage{booktabs}
\usepackage{wrapfig}
\usepackage{graphicx}
\usepackage{cite}
\usepackage[T1]{fontenc}
\usepackage{amsmath,amssymb}

\usepackage{amsfonts}

\usepackage{hyperref}
\usepackage{url}

\usepackage{booktabs}
\usepackage{nicefrac}

\usepackage{xcolor}

\usepackage{algorithm}
\usepackage{algpseudocode}

\usepackage{multirow}
\usepackage{array}

\usepackage{tikz}

\usepackage[inline]{enumitem}

\usepackage[us,12hr]{datetime}

\usepackage{newfloat}

\usepackage{caption}

\usepackage{tcolorbox}

\usepackage{fancyvrb}
\usepackage{bibentry}
\usepackage{listings}
\lstdefinelanguage{lean}{
  morekeywords={
    theorem,
  },
  sensitive=true,
  morecomment=[l]--,
  morestring=[b]",
}
\lstdefinelanguage{KB}{
  morecomment=[f][\color{gray}]{@@},
  morecomment=[f][\color{gray}]{diff},
  morecomment=[f][\color{gray}]{index},
  morecomment=[f][\color{gray}]{---},
  morecomment=[f][\color{gray}]{+++},
  morecomment=[f][\color{red}]{-},
  morecomment=[f][\color{green!70!black}]{+},
  morestring=[b]",
}
\definecolor{promptgray}{RGB}{245,245,245}
\definecolor{promptborder}{RGB}{200,200,200}
\newtcolorbox{promptbox}[1][]{
    colback=promptgray,
    colframe=promptborder,
    boxrule=1pt,
    arc=3pt,
    left=8pt,
    right=8pt,
    top=8pt,
    bottom=8pt,
    breakable,
    enhanced,
    #1
}
\newcommand{\method}{$\textsc{BREW}$}
\newcommand{\benchosworld}{\textsc{OSWorld}}
\newcommand{\benchtau}{$\tau^2$\text{-Bench}}
\newcommand{\benchss}{\textsc{SpreadsheetBench}}
\newcommand{\ignore}[1]{}
\usepackage{lineno}

\definecolor{darkblue}{rgb}{0, 0, 0.5}
\hypersetup{colorlinks=true, citecolor=darkblue, linkcolor=darkblue, urlcolor=darkblue}

\title{Improving Language Agents through \textsc{BREW}: Bootstrapping expeRientially-learned Environmental knoWledge}

\author{Shashank Kirtania$^{1}$\thanks{$^{1,2,4}$ Work done at Microsoft.\\Corresponding author: \texttt{priyansgupta@microsoft.com}}
\quad
Param Biyani$^{2}$
\quad
Priyanshu Gupta$^{3}$
\quad
Yasharth Bajpai$^{3}$
\quad
\\
Roshni Iyer$^{4}$
\quad
Sumit Gulwani$^{3}$
\quad
Gustavo Soares$^{3}$\\
\\
$^{1}$University of Michigan, \quad
$^{2}$Independent, \quad
$^{3}$Microsoft, \quad
$^{4}$Apple.
}

\begin{document}

\ifcolmsubmission
\linenumbers
\fi

\maketitle
\begin{abstract}
\label{sec:abstract}

\input{content_colm/abstract}
\end{abstract}

\section{Introduction}
\label{sec:introduction}

\input{content_colm/introduction}

\section{Related Works}
\label{sec:related-work}

\input{content_colm/related_work}

\section{BREW: Architecture}
\label{sec:architecture}
\input{content_colm/architecture}
\section{Experimental Setup}
\label{sec:experiments}
\input{content_colm/experiment}
\section{Analysis \& Discussion}
\label{sec:analysis}
\input{content_colm/analysis}
\section{Conclusions}
\label{sec:conclusion}
\input{content_colm/conclusion}

\newpage
\bibliographystyle{plainnat}
\bibliography{colm2026_conference}

\clearpage
\appendix
\input{content_colm/appendix_new}



\end{document}

%% file: content_colm/abstract.tex
Large Language Model (LLM)-based agents are increasingly capable of complex, multi-step tasks such as GUI automation, tool use, and data manipulation, yet they cannot learn from experience: each new session rediscovers solutions from scratch. We introduce \method~(\textbf{B}ootstrapping expe\textbf{R}ientially-learned \textbf{E}nvironmental kno\textbf{W}ledge), a framework that distills an agent's past interaction trajectories into a structured, retrievable knowledge base (KB) of natural-language \emph{recipes}, concept-level procedural documents that capture \emph{what} to do, \emph{when} it applies, and \emph{what} to watch out for. Drawing on the principle of library learning from program synthesis, \method~decomposes agent memory into modular, concept-localized documents and formalizes KB construction as a state-space search problem. To navigate this space, we introduce Expand-and-Gather Monte Carlo Tree Search (EG-MCTS), a reward-guided algorithm that jointly optimizes recipe accuracy and retrievability across parallel, per-concept search trees. We further adapt hindsight relabeling to convert near-miss trajectories into positive demonstrations, surfacing latent agent competencies as reusable knowledge. On three domain-grounded benchmarks, \benchosworld, \benchtau, and SpreadSheetBench, \method~achieves $10$--$20\%$ gains in task success and $10$--$15\%$ fewer execution steps over base agents, while consistently outperforming existing memory-augmented baselines that can degrade below memoryless performance. The resulting KB is inspectable, modular, and extensible, providing a transparent and controllable substrate for agent optimization.

%% file: content_colm/introduction.tex
Large Language Model (LLM) based agents are increasingly capable of interacting with complex environments like navigating GUIs, calling external tools, and manipulating structured data across a wide range of real-world tasks~\citep{li2025review, qin2025ui, swebench, yang2024swe, anthropic2024computer, openai2025operator}. Yet these agents lack a fundamental ability that humans take for granted: learning from experience. When an agent encounters a familiar task in a new session, it rediscovers the solution from scratch, repeating the same wrong menu paths, redundant API calls, or failed spreadsheet manipulations it had already resolved before.

After exporting a document to PDF once in LibreOffice, a person does not memorize the raw click sequence; they internalize a \emph{recipe}: the general pattern (File $\rightarrow$ Export as PDF), the conditions under which it applies (any Writer or Impress document), and the pitfalls to avoid (confirming the output opens correctly). Similarly, after handling a few customer return requests, a support agent learns the procedural structure: authenticate the user, verify the order status is \texttt{delivered}, collect all exchange items upfront because the step cannot be repeated, then execute. These recipes transfer across tasks within the same environment, compressing future problem-solving into a few reliable steps.

\textit{How should an agent acquire such recipes?} Weight optimization~\citep{schulman2017proximal, dpo2024, grpo2024} bakes experiential knowledge into model parameters, producing opaque policies that resist inspection or modular update. Memory-augmented agents~\citep{mem0, xu2025mem, gupta-etal-2024-metareflection, agrawal2025gepa} pursue explicit storage but at the wrong granularity, retaining either transient trajectory fragments or high-level reflections that lack the specificity to guide execution. A more principled template comes from library learning in program synthesis~\citep{ellis2020dreamcodergrowinggeneralizableinterpretable}, where agents refactor recurring subroutines into \emph{explicit, modular objects} that live outside the model, inspectable, updatable, and composable, dramatically compressing the search space for future tasks. We adopt the same principle but observe that the abstractions real-world agents need are not code primitives: they are \emph{natural-language recipes}, structured descriptions of what to do, when it applies, and what to watch out for, because the environments themselves (GUIs, tool APIs, conversational interfaces) are described and interacted with in language (see Section~\ref{sec:related-work} for a detailed discussion).

\begin{figure*}[t]
    \centering
    \includegraphics[width=\linewidth]{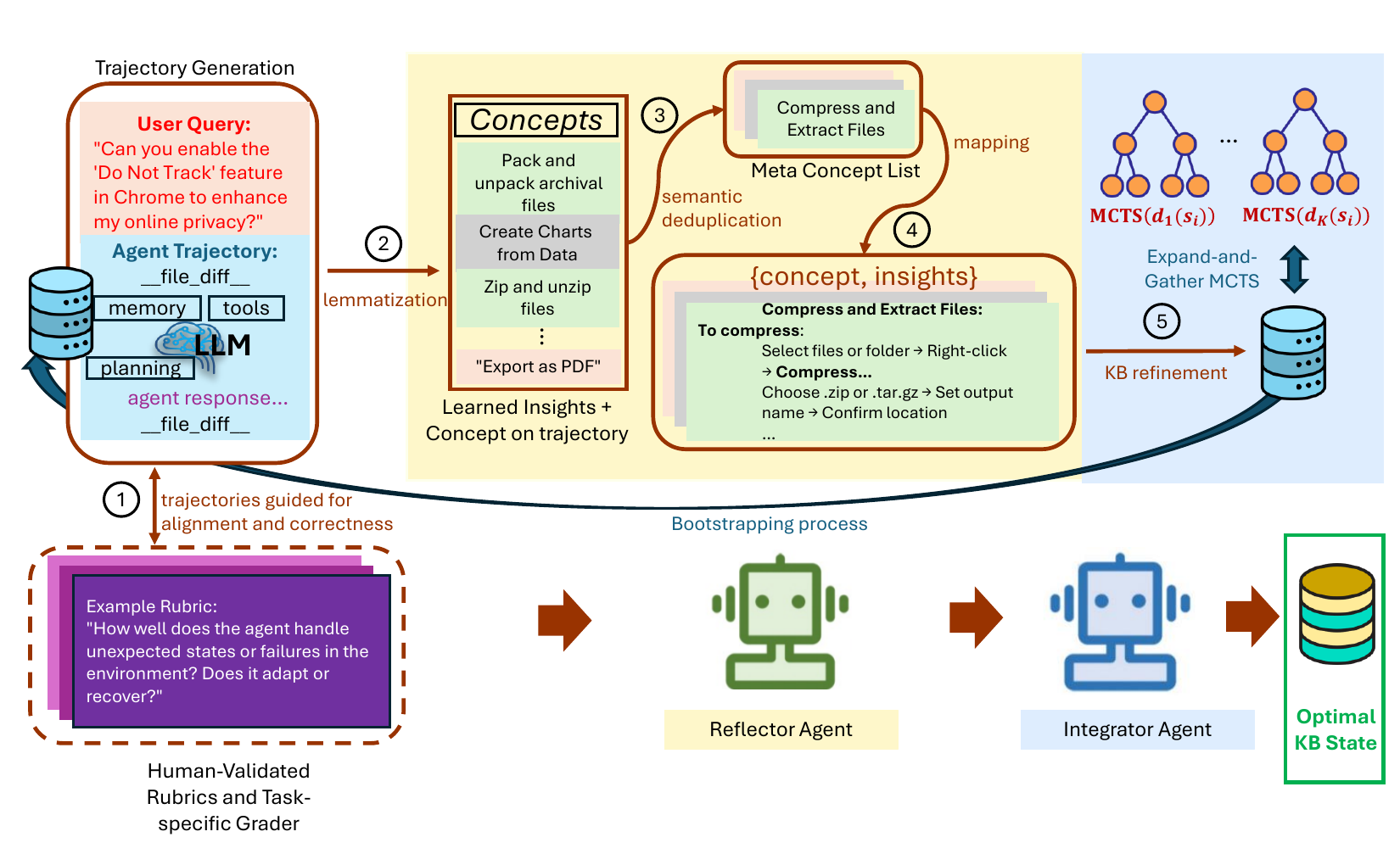}
    \caption{\textmd{\method~architecture overview using examples from the \benchosworld~dataset. Step 1 indicates trajectory generation with agent alignment to human-validated rubrics and correctness using a task-specific grader. Steps 2–4 indicate the Reflector Agent, which learns key concepts and insights from trajectories. Step 5 indicates the Integrator Agent, which integrates knowledge from the Reflector Agent to bootstrap the KB. We introduce Expand-and-Gather MCTS to find the best KB configuration by reward-guided search.}
    }
    \label{fig:architecture}
    \vspace{-3mm}
\end{figure*}

We introduce \method~(\textbf{B}ootstrapping expe\textbf{r}ientially-learned \textbf{e}nvironmental kno\textbf{w}ledge), a framework that learns a recipe library from agent experience (Figure~\ref{fig:architecture}). \method~incrementally constructs and refines a knowledge base (KB), a structured collection of concept-level procedural documents in natural language, directly from an agent's past interaction trajectories. Each document captures a reusable behavioral pattern at the right level of abstraction: more specific than a reflection (``be careful with file formats'') yet more general than a raw trajectory (the full click-by-click trace of one export). The KB serves as persistent, retrievable memory that the agent queries during future executions, compressing the search space the way a learned library compresses program synthesis. Because natural-language recipes lack the clean verification signals of compiled code, we formalize KB construction as a state-space search problem and introduce Expand-and-Gather Monte Carlo Tree Search (EG-MCTS), a reward-guided algorithm that finds recipe formulations that are both accurate for downstream reasoning and retrievable at inference time. Our key contributions are:

\begin{itemize}[leftmargin=*]
    \item \emph{Recipe learning from agent trajectories.} We propose a technique for distilling concept-level procedural recipes from agent interaction trajectories. Guided by rubrics and task-specific graders, this process partitions raw experience into modular, human-interpretable knowledge, the natural-language analog of library functions in program synthesis.

    \item \emph{State-space search for recipe optimization.} We formalize KB refinement as a state search problem and introduce EG-MCTS, which explores the space of recipe formulations via parallel, reward-guided search across concept-level documents, addressing the noise and ambiguity inherent in optimizing natural-language abstractions.

    \item \emph{Hindsight relabeling for broader recipe coverage.} We adapt hindsight relabeling~\citep{andrychowicz2018hindsightexperiencereplay} to expand the diversity of learned recipes, converting near-miss trajectories into positive demonstrations that reveal latent agent competencies as reusable knowledge.

    \item \emph{State-of-the-art results.} On domain-grounded benchmarks including \benchosworld~and \benchtau, \method~achieves $10$--$20\%$ gains in task precision and $10$--$15\%$ fewer execution steps, while maintaining memory and compute costs comparable to base LLMs.
\end{itemize}

%% file: content_colm/related_work.tex
\paragraph{Agent Learning from Demonstrations}  
The principle of \emph{library learning} from program synthesis---where agents refactor recurring subroutines into reusable primitives that compress future search~\citep{ellis2020dreamcodergrowinggeneralizableinterpretable}---motivates a growing body of work on structured skill extraction for LLM agents. Recent methods leverage LLMs to isolate reusable skills through interactive decomposition~\citep{hashemzadeh2024subgoaldistillationmethodimprove}, synthesize executable domain-specific functional abstractions~\citep{khan2025executable}, or learn in-prompt memory~\citep{gupta-etal-2024-metareflection}. However, these approaches rely on static decomposition or offline synthesis, and the code-level primitives they produce do not naturally capture procedural, environment-specific knowledge (e.g., that ``Ctrl+F opens search in Nautilus''). \method~addresses both limitations: it iteratively refines natural-language \emph{recipes} via rollout-generated insights and structured knowledge-base search (EG-MCTS). Unlike prompt-optimization techniques~\citep{agrawal2025gepa, gupta-etal-2024-metareflection}, the resulting recipes are stored as retrievable, extensible agent-memory knowledge bases.

\paragraph{Agentic Memory}
The concept of providing agents with controllable memory has a rich history~\citep{Littman1993AnOC}. Memory mechanisms are attracting increasing attention~\citep{packer2024memgptllmsoperatingsystems, wang2025lifespancognitivesystems, xu2025raginthewild, mem0, xu2025amemagenticmemoryllm, hu2025evaluatingmemoryllmagents}, typically storing relevant context in structured formats such as graphs or trees for retrieval-augmented generation. While effective in their target sub-domains, these techniques fail to generalize across environments~\citep{hu2025evaluatingmemoryllmagents}. More fundamentally, model weight optimization~\citep{schulman2017proximal, dpo2024, grpo2024} learns abstractions implicitly by baking them into parameters, producing opaque policies that resist inspection or modular update and couple tightly to training distributions. At the other extreme, memory-augmented agents such as Mem0~\citep{mem0, xu2025mem} store transient trajectory fragments that vanish between episodes, while MetaReflection~\citep{gupta-etal-2024-metareflection} and GEPA~\citep{agrawal2025gepa} embed high-level reflections that lack the specificity to guide execution.
\paragraph{State Based Explorations} State-space search has been extensively used for exploration based learning ~\citep{alphago,Liu2025AlphaGoMF}. With the advent of prompt-tuned LLM systems, state space techniques are being actively explored in the community for text-based optimization~\citep{stackfeed, wang2023promptagent, Novikov2025AlphaEvolveAC}. Our technique builds upon this work and generalizes it to general-purpose agent memory learning.

%% file: content_colm/architecture.tex
\vspace{-0.7pt}
This section describes our proposed {\textbf{B}ootstrapping}
{expe\textbf{R}ientially-learned } {\textbf{E}nvironmental}
{kno\textbf{W}ledge} technique, \method, which constructs and iteratively refines a Memory KB using trajectory insights guided by human-validated general-purpose agent behavior metrics, task-specific evaluation, and latent insight generation. We decompose the problem of learning the optimal KB by partitioning memory as local documents associated with semantic concepts, and solve the KB learning problem by our novel Expand-and-Gather Monte Carlo Tree Search (EG-MCTS) algorithm. Figure~\ref{fig:architecture} provides an architecture overview of \method, and Algorithm~\ref{algo:overview} in Appendix~\ref{app:brewdetails} describes the full pseudocode.

\subsection{Trajectory Generation} Given the training dataset, we generate full-length trajectories, hereby referred to as rollouts, for each query using an LLM-powered agent 
conditioned on its associated KB. At initialization, the KB is empty, and we generate rollouts with an \emph{empty} KB.  Each rollout is evaluated using a 
correctness grader, which assigns a binary success label
and an LLM based qualitative assessment against a set of human-validated general-purpose agent behavior rubrics ~\citep{biyani2024RUBICON} (Step 1 in Figure~\ref{fig:architecture}). To enrich the diversity and coverage of these trajectories, we additionally apply hindsight relabeling (Section~\ref{sec:hindsight_method}).

\subsection{Hindsight Relabeling}
\label{sec:hindsight_method}
To expand the diversity of training signal beyond the original rollouts, we apply hindsight relabeling~\citep{andrychowicz2018hindsightexperiencereplay}. We sample failing trajectories and prompt an LLM to generate an alternative task description under which the agent's observed behavior would constitute a valid solution. Re-executing the agent on this reframed task frequently succeeds, converting roughly half of the originally failing trajectories into passing ones and shifting the effective distribution.

\subsection{Reflector and Integrator Agents}
\label{sec:reflector_integrator}
\paragraph{Reflector Agent:} 
\texttt{ReflAgent} takes as input a rollout with its rubric and correctness labels, and outputs sentence-level insights with mapped concepts:
\begin{equation}
\label{eq:reflector}
\{\textit{concepts}, \textit{insights}\} = \texttt{ReflAgent}(\{\textit{rollout}, \textit{eval}\}).
\end{equation}
Examples of concept\textendash insight pairs appear in Step~2 of Figure~\ref{fig:architecture}.  

\paragraph{Concept Deduplication:}  
Concept\textendash insight pairs are annotated independently per rollout, often producing overlapping or paraphrased concepts. We address this via semantic clustering (Steps~3\textendash 4, Figure~\ref{fig:architecture}; Algorithm~\ref{algo:overview}, line~3 in Appendix~\ref{app:brewdetails}): contextual embeddings for each concept are generated using an LLM, clustered, and each insight is mapped to its cluster representative. Details appear in Algorithms~\ref{alg:generateinsights} and \ref{alg:deduplicateconcepts} in Appendix~\ref{app:brewdetails}.

\paragraph{Integrator Agent:}  
\texttt{IntegAgent} incrementally builds and refines KB documents $\{d(s_i)\} \in \mathcal{D}(s_i)$ during environment interaction. Instead of a centralized memory, the KB is partitioned into local documents, each tied to a meta concept. This design enables (1) efficient, context-specific retrieval; (2) modular updates with minimal interference; and (3) natural alignment with task semantics, as deduplicated meta concepts capture meaningful behavioral abstractions. Unlike prior work assuming flat memory or dialogue histories, this structure is well-suited for long-horizon, procedural tasks where behaviors cluster around discrete skills.  

The KB is dynamically populated: concepts central to the dataset receive more updates. At each state, for meta concept $k$, \texttt{IntegAgent} updates its document via $d_k(s_{i+1}) \leftarrow \texttt{IntegAgent}(k, \textit{insights}_{k}, d_k(s_i))$, so the full KB is $\mathcal{D}(s_i) = \bigcup_{k \in \mathcal{K}} \{ d_k(s_i) \}$. To reduce LLM variance we use Expand-and-Gather MCTS (EG-MCTS; Figure~\ref{fig:kb_search}).

\subsection{Expand-and-Gather MCTS for Optimal KB Search}
\label{mcts}

We initialize the meta-concept set $\mathcal{K}$ by deduplicating concepts extracted by \texttt{ReflAgent} from the first set of rollouts, then create an empty document per concept $k \in \mathcal{K}$.

We model finding $\mathcal{D}^*$ as a search over the state space of all possible KBs. By decomposing $\mathcal{D}$ into concept-level documents, the problem reduces to finding $d_k^*$ for each concept $k$, with the full KB assembled as $\mathcal{D}^* = \bigcup_{\forall k}\{d_k^*\}$.
However, documents are \emph{not} independent: an agent can retrieve any document during inference, making it difficult to assess changes in isolation. Expand-and-Gather MCTS (EG-MCTS) addresses this by searching the per-concept state spaces concurrently via parallel MCTS explorations synced after each iteration. Node expansions proceed independently, but reward calculation and insight generation are conditioned on a running-optimum KB state. Each iteration consists of two phases:

\begin{figure*}[t]
    \centering
    \includegraphics[width=\linewidth]{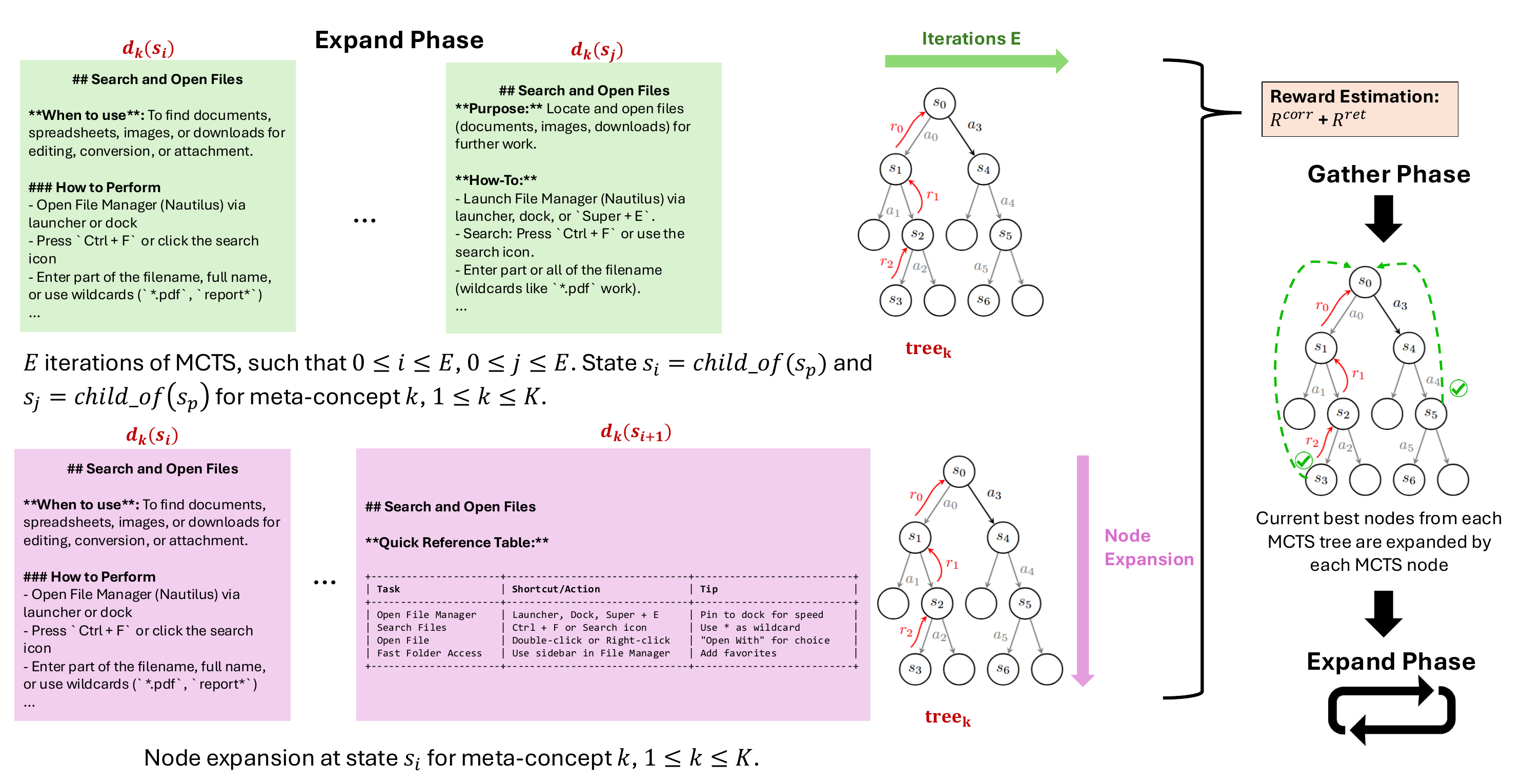}
    \caption{textmd{Illustration of \method's KB optimization process using Expand-and-Gather MCTS with \benchosworld~examples. In the \textbf{Expand Phase}, for each document $k$, we sample the best node from tree$_k$ using UCT and perform node expansion. Node rewards are estimated based on correctness and retrievability. In the \textbf{Gather Phase}, the best nodes from each tree are gathered at each node. The process is repeated for the next iteration of KB refinement.}
    }
    \label{fig:kb_search}
    \vspace{-3mm}
\end{figure*}

\paragraph{Expand Phase:} For each search tree, we select the best state $s^*$ and expand it concurrently. The expansion KB combines the current document $d_k(s^*)$ with the best documents for all other concepts: $\mathcal{D}_t = \{d_t\} \cup \{d^*_{i : i \neq t}\}$. Rollouts generated under $\mathcal{D}_t$ are consumed by \texttt{ReflAgent} for insights, after which \texttt{IntegAgent} produces updated document variants whose rewards are estimated and back-propagated through the tree.

\paragraph{Gather Phase:} The current best states from each document's MCTS tree are \textit{gathered} and distributed to every tree, enabling (i) reward estimation for each expanded state and (ii) new insight generation for further expansion.

\subsection{Reward-Guided Optimization} 
\paragraph{Reward Objective:}
Each document state is rewarded based on two complementary criteria: \emph{(i)} how well the current document contributes to \textbf{accurate downstream reasoning}, and \emph{(ii)} how \textbf{retrievable} it is in the context of a growing KB. Formally, the total reward at time step $t$ is defined as:
\begin{equation}
R_t = \lambda_{\text{corr}} \cdot R^{\text{corr}}_t + \lambda_{\text{ret}} \cdot R^{\text{ret}}_t
\end{equation}
where $R^{\text{corr}}_t$ is the \textbf{correctness reward}, $R^{\text{ret}}_t$ is the \textbf{retrieval reward}, and $\lambda_{\text{corr}}, \lambda_{\text{ret}} \in [0, 1]$ are scalar weights with $\lambda_{\text{corr}} + \lambda_{\text{ret}} = 1$.
\paragraph{Correctness Reward:}
The correctness reward $R^{\text{corr}}_t$ evaluates the accuracy of the agent's output over a held-out query set $\mathcal{Q}$, when reasoning over the current KB $\mathcal{D}_t$. It is defined as:
\begin{equation}
R^{\text{corr}}(d_t|\mathcal{D}_t) = \frac{1}{|\mathcal{Q}|} \sum_{q \in \mathcal{Q}} \text{Eval}_{\text{task}}(q, \texttt{agent} \oplus \mathcal{D}_t)
\end{equation}
where $\text{Eval}_{\text{task}}$ is a task-specific evaluation function (e.g., question-answering accuracy, entailment correctness), and $\texttt{agent} \oplus \mathcal{D}_t$ denotes the agent acting over the hybrid KB.
\paragraph{Retrieval Reward:}
The retrieval reward $R^{\text{ret}}_t$ measures how effectively the current document $d_t$ can be retrieved from the current KB $\mathcal{D}_t$. For a held-out query set $\mathcal{Q}$, it is computed using the mean reciprocal rank (MRR):
\begin{equation}
R^{\text{ret}}(d_t | \mathcal{D}_t) = \frac{1}{|\mathcal{Q}|} \sum_{q \in \mathcal{Q}} \text{MRR}_q(d_t, \mathcal{D}_t)
\end{equation}



\subsection{Complexity Analysis}
The overall time complexity of \method~is $O(|\mathcal{Q}_{\text{train}}| \cdot T_{\text{LLM}} + M \cdot |\mathcal{K}| \cdot h \cdot T_{\text{agent}})$, where $|\mathcal{Q}_{\text{train}}|$ is the number of training queries, $T_{\text{LLM}}$ is the cost of a single LLM call, $M$ is the number of MCTS iterations, $|\mathcal{K}|$ is the number of meta concepts, $h$ is the number of candidates per expansion, and $T_{\text{agent}}$ is the cost of a single agent rollout. See Algorithm~\ref{algo:overview} in Appendix~\ref{app:brewdetails} for the pseudocode and details about compute analysis in Appendix \ref{app:compute_cost} .

%% file: content_colm/experiment.tex
\paragraph{Datasets}

We evaluate \method~on three benchmarks testing different interactive agent capabilities: \textsc{OSWorld} for computer-use automation~\citep{osworld}, $\tau^2$-Bench for tool use~\citep{barres2025tau2benchevaluatingconversationalagents}, and \textsc{SpreadsheetBench} for data manipulation~\citep{spreadsheetbench}. 
\begin{enumerate}[leftmargin=*]
    \item \textbf{OSWorld:} This benchmark tests multimodal agents on real-world computer tasks across 10 applications. We use \textit{GTA1-7B}, a state-of-the-art computer-use agent with \method. We split the data into 20 train and 349 test samples, balanced across application categories.
Tasks are evaluated using 134 custom scripts that verify final application states.

    \item \textbf{$\boldsymbol{\tau^2}$-Bench:} This benchmark evaluates conversational agents on multi-turn tool-use scenarios across \emph{Telecom}, \emph{Retail}, and \emph{Airline} domains. We test \texttt{o4-mini}-based tool-calling agent. We use 20 train and 254 test samples, balanced across domains.
    
\item \textbf{SpreadsheetBench:} This benchmark evaluates agents on real-world spreadsheet manipulation, spanning both cell-level and sheet-level tasks. It contains 912 authentic user instructions paired with 2,729 test cases (3 per instruction).
We test \texttt{o4-mini} using a Python tool-calling agent, and enhance it with an embedding based Retrieval over the \method~KB generated over a small held-out train set of 30 samples.

\end{enumerate}

\paragraph{Agents \& KB Integration} For \benchosworld, we augment \textit{GTA1-7B}~\citep{yang2025gta1guitesttimescaling} (paired with O3 or GPT5.2 as grounding model). Retrieved documents are injected into the planner's system prompt; for the grounding agent, retrieval is performed at each grounding step and provided in the user prompt. For \benchtau~and \benchss, we augment tool-calling agents by prepending the top retrieved entries to the system prompt.

\paragraph{Baselines} We compare \method~against two widely used experiential memory approaches, \textit{Cognee}\footnote{\href{https://github.com/topoteretes/cognee}{github.com/topoteretes/cognee}}~\citep{markovic2025optimizinginterfaceknowledgegraphs} and \textit{Agent-Mem}~\citep{xu2025amemagenticmemoryllm}, both of which serve as established baselines for AI memory evaluation.
Cognee uses a graph-plus-vector memory architecture to build cross-document connections from trajectories, while Agent-Mem provides a scalable memory layer for extracting and retrieving information from conversational data with graph-based representations.

\paragraph{Other Experimental Configs:} For all experiments, expansion width $e=3$, max depth $k=3$, and balanced reward weights $\lambda_{\text{corr}} = \lambda_{\text{ret}} = 0.5$. During MCTS node selection, we use the UCT~\citep{uct} for balancing exploration and exploitation. Full experimental details are provided in the Appendix.

%% file: content_colm/analysis.tex
In this section, we present findings from our evaluation of \method. For more details on qualitative insights and discussion you may refer to the supplementary material. 
\vspace{-0.1cm}
\subsection{Main Results}
Table~\ref{tab:main_comparison} summarizes performance across all three benchmarks. \method~outperforms every baseline on every benchmark, achieving absolute gains of +8.4 on \benchosworld~(GTA1-7B+o3), +5.3 (GTA1-7B+GPT5.2), +5.87 on \benchtau, and +4.2 on \benchss. Notably, both Cognee and Agent-Mem sometimes \emph{degrade} performance below the no-memory baseline (e.g., Agent-Mem on \benchosworld: 43.83 vs.\ 53.30; Cognee on \benchss: 42.10 vs.\ 44.30), suggesting that na\"ive memory injection can hurt when retrieved context is misaligned. \method~avoids this failure mode through reward-driven state search, which optimizes KB content before deployment rather than injecting raw trajectory fragments at inference time.
\vspace{-0.3cm}
\subsection{Variations Across State Search Strategy}
\label{sec:search_strategy} \method~treats KB construction as a search problem over possible KB states. We compare MCTS against \textit{Iterative Refinement} and \textit{Greedy Search} (details in Appendix~\ref{app:search_strategies}). Table~\ref{tab:ablation} presents results across all three benchmarks.
\vspace{-0.3cm}
\paragraph{Iterative refinement is fragile under LLM stochasticity.}
It degrades below baseline on \benchss~(42.98 vs.\ 44.30) and barely
matches it on \benchosworld~(53.10 vs.\ 53.30), with only a marginal gain on \benchtau~(57.34 vs.\ 56.63)~(Table~\ref{tab:ablation}).
Without branching, a single bad refinement step has no recovery mechanism.
\vspace{-0.3cm}
\paragraph{Greedy search suffices for simple domains; MCTS excels on complex ones.}
On \benchosworld, MCTS outperforms greedy by 3.7 points (56.80 vs.\ 53.10),
while on \benchtau~and \benchss~the gap narrows (Table~\ref{tab:ablation}), consistent with \benchosworld's richer inter-document dependencies benefiting from broader exploration.
\vspace{-0.1cm}

\begin{table*}[]
\vspace{-1cm}
\centering
\resizebox{\linewidth}{!}{%
\begin{tabular}{lcccc}
\textbf{Method} & \textbf{OSWorld Verified} & \textbf{OSWorld Verified} & \textbf{$\tau^2$ Bench} & \textbf{SpreadsheetBench} \\ 
 & GTA1-7B + o3 & GTA1-7B + GPT5.2 & o4-mini & o4-mini\\ \midrule
Baseline       & 53.30 & 63.04 & 56.63 & 44.30 \\
Cognee         & 46.70 & 58.55 & 57.71 & 42.10 \\
Agent-Mem      & 43.83 & 57.28 & 52.69 & 42.00 \\ \midrule
\method~ & \textbf{61.70} & \textbf{68.34} & \textbf{62.50} & \textbf{48.50} \\ 
\bottomrule
\end{tabular}%
}
\caption{Comparison against baselines -- task success rate, ratio of independent successes, and first-testcase pass rate for OSWorld Verified, $\tau^2$ Bench, and SpreadsheetBench, respectively.}
\label{tab:main_comparison}
\end{table*}


\subsection{\method~learns recipes from sub-trajectories in \benchosworld.}
 
\begin{figure}[t]
    \centering
    \includegraphics[width=0.9\columnwidth]{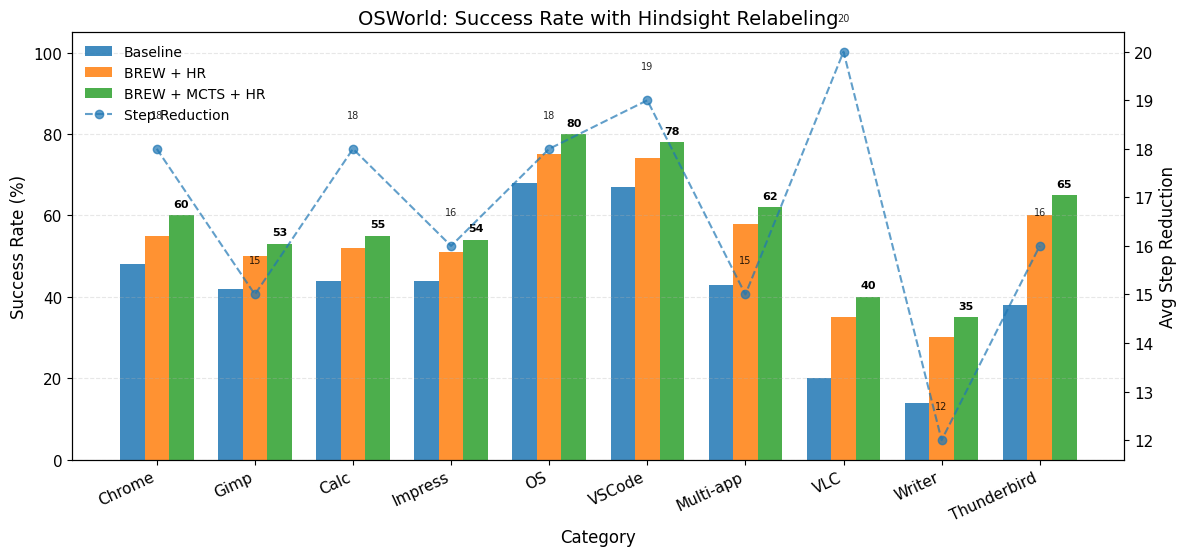}
    \caption{The bar plot represents the category-wise success rate over various tasks in the \benchosworld~dataset over the GTA1-agent, whereas the line plot demonstrates the reduction in the number of steps for the successful cases. Note that even in scenarios where the KB doesn't help increase the success rate, it steeply reduces the \#steps needed to succeed.}
    \label{fig:osworld_success_compl}
    \vspace{-0.5cm}
\end{figure}

Figure~\ref{fig:osworld_success_compl} breaks down performance by OSWorld category. The OSWorld analysis identifies three principal failure modes for agents: GUI grounding, operational knowledge, and long-horizon planning. \method's KB directly addresses the second by distilling procedural recipes from trajectories.
 
\textbf{Accuracy gains track operational-knowledge bottlenecks.} Writer and Thunderbird, which rely on repetitive menu-driven workflows with reusable sub-procedures, show the strongest accuracy improvements. VSCode benefits similarly from distilled IDE command sequences. GIMP and Calc, by contrast, see negligible accuracy change: GIMP tasks are bottlenecked by fine-grained spatial manipulation, while Calc tasks demand per-instance formula logic with little cross-task procedural overlap.
 
\textbf{Efficiency decouples from correctness.} The step-reduction signal (dashed line) reveals that even in categories where accuracy is flat, \method~reduces execution length by 14--23 steps. The KB is retrieved and followed for the operational-knowledge component, but the task ultimately fails at a downstream bottleneck the KB cannot address, such as a grounding misclick or a missing perceptual inference. This decoupling indicates that future progress in these categories requires complementing procedural memory with improved visual grounding rather than richer KBs.

\subsection{\method~learns aggressive resolution strategies for \benchtau}
Table~\ref{tab:tau_errs}, breaks down error types on \benchtau~Retail. Prior failure analysis shows that \textit{Wrong info} and \textit{Wrong decision} errors are predominantly terminal: once the agent provides a miscalculated price or violates a policy constraint (e.g., calling exchange tools item-by-item when a batched call is required), the conversation diverges irrecoverably. \method~targets this by surfacing verified policy resolutions at retrieval time, preventing the initial misstep rather than attempting recovery. Due to this \textbf{\method~prevents terminal policy errors by front-loading verified resolutions at retrieval time}. Other baselines occasionally increase errors because $\tau$-bench policies contain subtle conditional distinctions: return eligibility depends on order status, item category, and time window, and retrieving a precedent matching request type but differing on one condition injects a wrong constraint.
 
\subsection{BREW learns domain-specific strategies for SpreadsheetBench}

\textbf{Domain-specific KB entries correct recurring baseline failure modes.} \method~shows consistent improvements over the Baseline on SpreadsheetBench. We identified 109 cases where \method~succeeded on the first test case while the baseline failed, with identical inputs aside from KB augmentation. In most of these cases the difference traced to more precise formula placement, followed by result verification before submission and correct filter usage. Three failure modes account for the majority of corrected errors: (i) header misdetection, where the baseline promotes the first data row to column names in headerless spreadsheets; (ii) fragile formula construction, such as using string comparison instead of ISBLANK for blank-cell detection; and (iii) scope violations, where baseline agents write beyond the specified cell range or substitute Python lookups for requested Excel formulas.

\textbf{KB-guided validation prevents cascading errors.} The Header Extraction entry guides the agent to validate column metadata before processing, while the Writeback Results and Column Selection entries enforce range compliance and appropriate formula idioms. Beyond correctness, \method-augmented trajectories consistently include verification steps such as diff generation, output reloading, and formula spot-checks, surfaced through the Difference in State KB entry. These routines catch errors early in multi-step manipulation chains that would otherwise propagate silently. We provide full case studies in Appendix~\ref{ssbench_anal}.
\begin{table*}[]
\centering
\resizebox{\linewidth}{!}{%
\begin{tabular}{lcccc}
\toprule
\textbf{Method} & \textbf{OSWorld Verified} & \textbf{OSWorld Verified} & \textbf{$\tau^2$ Bench} & \textbf{SpreadsheetBench} \\ 
 & GTA1-7B + o3 & GTA1-7B + GPT5.2 & o4-mini & o4-mini\\ \midrule
\method~-Iterative & 59.40 & 67.29 & 59.14 & 47.13 \\
\method~-Greedy & 59.40 & 67.29 & 60.71 & 46.66 \\
\method~-MCTS& \textbf{61.70} & \textbf{68.34} & \textbf{62.50} & \textbf{48.50} \\  
\bottomrule
\end{tabular}%
}
\caption{Ablation over \method~ variants. We compare search strategies (Iterative, Greedy, MCTS) and the effect of Hindsight Relabeling across all three benchmarks.}
\label{tab:ablation}
\end{table*}

\subsection{Effect of Hindsight Relabeling}
\label{sec:hindsight}
Hindsight relabeling consistently improves task success rates across all benchmarks and search strategies (Table~\ref{tab:ablation}).
The key insight is that the same agent behavior receives opposite polarity from the Reflector depending on whether the trajectory is labeled as passing or failing (Eq.~\ref{eq:reflector}). A failing trajectory produces diagnostic insights (``what went wrong''), while the same behavioral pattern, relabeled as passing under a new task, produces prescriptive insights (``what to do''). This dual signal enriches the KB with both cautionary and actionable procedural knowledge. Additionally, re-execution on the relabeled task is not guaranteed to reproduce the original trajectory exactly, exposing nearby behavioral modes that give the Reflector genuinely novel material.

The resulting shift from 50/50 to roughly 66\% passing / 33\% failing biases KB construction toward diverse successful demonstrations, representing latent competences never surfaced as successes in the original data. While retaining enough failure coverage to prevent overconfidence. See Appendix \ref{app:hindsight_details} for the full analysis.
\vspace{-2.5pt}
\subsection{Knowledge Base Fidelity on Training Trajectories}

Table~\ref{tab:train_acc} reports train-set accuracy as a proxy for how faithfully each method's KB encodes knowledge from its construction trajectories. All memory-augmented methods outperform the memoryless baseline on the train split, but cross-referencing with test results (Table~\ref{tab:main_comparison}) reveals that high train fidelity can be actively harmful. Agent-Mem matches \method on $\tau^2$-Bench train yet collapses to 52.69 on test, below the memoryless baseline. Cognee shows the same inversion on OSWorld. We find Agent-Mem stores near-verbatim episodic traces, so retrieval at test time surfaces action sequences tightly coupled to the original task's state transitions; minor shifts in goal or initial state turn the retrieved trace into a misleading prior. Cognee's entity-centric graph retains relational structure but loses procedural ordering, reconstructing what entities were involved but not \textit{in what sequence} or \textit{under what conditions}. Both pathologies yield negative transfer: the KB retrieves confidently but incorrectly.
\method avoids this because the Reflector distills trajectories into condition-action rules indexed by environment state rather than by source task identity. Retrieval therefore generalizes across tasks sharing procedural sub-structure even when surface-level goals differ, making \method the only method where train fidelity consistently converts to test gains across all three benchmarks.
\vspace{-3.0pt}

%% file: content_colm/conclusion.tex
\vspace{-4.5pt}We introduced \textsc{BREW}, a framework that constructs and refines a structured, interpretable knowledge base from past agent interactions, offering an alternative to direct model fine-tuning. By decomposing agent memory into concept-level documents and optimizing their content via state-space search, \textsc{BREW} provides a modular and transparent substrate for memory formation. Evaluations across \benchosworld, \benchtau, and \benchss~show that structured memory yields measurable improvements in task success and efficiency at manageable computational cost. Although promising, \textsc{BREW}'s effectiveness depends on training data quality and coverage. Future work could explore more adaptive, domain-general refinement techniques and tighter integration with agent planning. We hope this study encourages further investigation into interpretable, memory-driven approaches to language agent development, particularly in real-world environments where long-term consistency and adaptability are essential.

%% file: content_colm/appendix_new.tex
\section{Appendix}
\label{app:brewdetails}

\subsection{Details of the \method~Algorithm}

We provide pseudocode for the core components of \method, aligning with the stages introduced in Section~\ref{sec:architecture}. Algorithm~\ref{algo:overview} describes the overall \method~procedure. \textsc{GenerateInsights} (Alg.~\ref{alg:generateinsights}) produces concept-aligned insights from annotated rollouts using \texttt{ReflAgent}. \textsc{DeduplicateConcepts} (Alg.~\ref{alg:deduplicateconcepts}) clusters semantically overlapping concepts into a compact meta-concept set. \textsc{IntegAgent}  incrementally builds and updates per-concept documents using newly generated insights. Finally, \textsc{ExpandNode} (Alg.~\ref{alg:expandnode}) performs MCTS-guided expansions to explore improved document variants, while \textsc{Evaluate} (Alg.~\ref{alg:evaluate}) scores candidate KB states using correctness and retrieval-based rewards.

\begin{algorithm}[H]
\caption{BREW: Bootstrapping Experientially-learned Environmental Knowledge}
\begin{algorithmic}[1]
\Require Training samples $\mathcal{Q}_{\text{train}}$, eval samples $\mathcal{Q}_{\text{eval}}$, rubrics, iterations $M$, candidates per expansion $h$
\Ensure Optimized KB $\mathcal{D}^*$

\vspace{3pt}
\Statex \textbf{Initialization}
\vspace{2pt}

\State $\mathcal{D}_0 \gets \varnothing$
\State $\mathcal{B} \gets \textsc{GenerateInsights}(\mathcal{Q}_{\text{train}}, \mathcal{D}_0, \text{rubrics})$
\State $\mathcal{B} \gets \mathcal{B} \cup \textsc{HindsightRelabel}(\mathcal{Q}_{\text{train}}, \mathcal{D}_0, \text{rubrics})$ \Comment{Relabel failing trajectories}
\State $\mathcal{K} \gets \textsc{DeduplicateConcepts}(\mathcal{B})$ \Comment{Initial concept set}
\For{each $k \in \mathcal{K}$}
    \State $d_k^0 \gets \textsc{IntegAgent}(k, \mathcal{I}_k, \varnothing)$
    \State Initialize $\text{tree}_k$ with root node $d_k^0$
\EndFor
\State $\mathcal{D}_{\text{current}} \gets \bigcup_{k \in \mathcal{K}} \{d_k^0\}$ \Comment{Initial KB}

\vspace{3pt}
\Statex \textbf{EG-MCTS Optimization}
\vspace{2pt}

\For{$t = 1$ to $M$}
    \Comment{Parallel expansion across concepts}
    \For{each $k \in \mathcal{K}$}
        \State $s_k \gets \textsc{SelectBestNode}(\text{tree}_k)$ \Comment{UCT selection}
        \State $\mathcal{D}_{\text{best}} \gets \bigcup_{k' \in \mathcal{K}} \{d_{k'}^{\text{best}}\}$ \Comment{Current best docs}
        \State \textsc{ExpandNode}($s_k$, $k$, $h$, $\mathcal{D}_{\text{current}}$, $\mathcal{D}_{\text{best}}$, $\text{tree}_k$)
    \EndFor
    \Comment{Update current best documents}
    \For{each $k \in \mathcal{K}$}
        \State $d_k^{\text{best}} \gets$ best document in $\text{tree}_k$
    \EndFor
    \State $\mathcal{D}_{\text{current}} \gets \bigcup_{k \in \mathcal{K}} \{d_k^{\text{best}}\}$
\EndFor
\State \Return $\mathcal{D}_{\text{current}}$
\end{algorithmic}
\vspace{6pt}
\label{algo:overview}

\textbf{Time Complexity:} $O(|\mathcal{Q}_{\text{train}}| \cdot T_{\text{LLM}} + M \cdot |\mathcal{K}| \cdot h \cdot T_{\text{agent}})$
\end{algorithm}

\begin{algorithm}[H]
\caption{GenerateInsights: Extract behavioral insights from trajectories}
\label{alg:generateinsights}
\begin{algorithmic}[1]
\Require Queries $\mathcal{Q}$, KB $\mathcal{D}$, rubrics
\Ensure Concept-insight pairs $\mathcal{B}$

\State $\mathcal{B} \leftarrow \varnothing$
\For{each query $q \in \mathcal{Q}$}
    \State $\tau \leftarrow$ \Call{LLM}{$q, \mathcal{D}$} \Comment{Generate trajectory}
    \State label $\leftarrow$ \Call{Grade}{$\tau$} \Comment{Success/failure}
    \State $(c, i) \leftarrow$ \Call{ReflAgent}{$\tau, \text{rubrics}, \text{label}$}
    \State $\mathcal{B} \leftarrow \mathcal{B} \cup \{(c, i, q)\}$ \Comment{Store with source query}
\EndFor
\State \Return $\mathcal{B}$
\end{algorithmic}
\end{algorithm}

\begin{algorithm}[H]
\caption{DeduplicateConcepts: Cluster similar concepts and map queries}
\label{alg:deduplicateconcepts}
\begin{algorithmic}[1]
\Require Concept-insight-query triples $\mathcal{B}$
\Ensure Meta-concepts $\mathcal{K}$ with mapped queries and insights

\State Extract all concepts from $\mathcal{B}$
\State Embed and cluster concepts by similarity
\State $\mathcal{K} \leftarrow$ cluster representatives
\For{each $k \in \mathcal{K}$}
    \State $\mathcal{Q}_k^{\text{train}} \leftarrow$ \{training queries that contributed insights to $k$\}
    \State $\mathcal{Q}_k^{\text{eval}} \leftarrow$ \{held-out queries relevant to $k$\}
    \State $\mathcal{I}_k \leftarrow$ \{all insights mapped to concept $k$\}
\EndFor
\State \Return $\mathcal{K}$ with associated queries and insights
\end{algorithmic}
\end{algorithm}

\begin{algorithm}[H]
\caption{ExpandNode: Generate and evaluate new document variants}
\label{alg:expandnode}
\begin{algorithmic}[1]
\Require Node $s$, concept $k$, candidates $h$, current KB $\mathcal{D}_{\text{current}}$, best docs $\mathcal{D}_{\text{best}}$, tree
\Ensure Updated tree with new evaluated nodes

\State \Comment{Generate new insights from concept-relevant queries}
\State $\mathcal{B}_{\text{new}} \leftarrow \varnothing$
\For{query $q \in \mathcal{Q}_k^{\text{train}}$}
    \State $\tau \leftarrow$ \Call{LLM}{$q, \mathcal{D}_{\text{current}}$}
    \State $(c, i) \leftarrow$ \Call{Annotate}{$\tau, \text{rubrics}, \cdot$}
    \If{$c$ maps to $k$}
        \State $\mathcal{B}_{\text{new}} \leftarrow \mathcal{B}_{\text{new}} \cup \{i\}$
    \EndIf
\EndFor

\State \Comment{Generate and evaluate candidate documents}
\For{$j = 1$ to $h$}
    \State $d_{k,j} \leftarrow$ \Call{IntegAgent}{$k, \mathcal{I}_k \cup \mathcal{B}_{\text{new}}, d_k^s$}
    
    \State \Comment{Evaluate using hybrid KB with best docs from other concepts}
    \State $\mathcal{D}_{\text{hybrid}} \leftarrow \{d_{k,j}\} \cup \{d_{k'} \in \mathcal{D}_{\text{best}} : k' \neq k\}$
    \State $R_{k,j} \leftarrow$ \Call{Evaluate}{$d_{k,j}, \mathcal{D}_{\text{hybrid}}, \mathcal{Q}_k^{\text{eval}}$}
    
    \State \Comment{Add to tree and backpropagate}
    \State Add $(d_{k,j}, R_{k,j})$ as child of $s$ in tree
    \State Backpropagate $R_{k,j}$ from new node to root
\EndFor
\end{algorithmic}
\end{algorithm}

\begin{algorithm}[H]
\caption{Evaluate: Score document using held-out queries}
\label{alg:evaluate}
\begin{algorithmic}[1]
\Require Document $d_k$, hybrid KB $\mathcal{D}_{\text{hybrid}}$, eval queries $\mathcal{Q}_k^{\text{eval}}$
\Ensure Reward score $R$

\State $R^{\text{corr}} \leftarrow 0$
\State $R^{\text{ret}} \leftarrow 0$
\For{each $q \in \mathcal{Q}_k^{\text{eval}}$}
    \State $R^{\text{corr}} \leftarrow R^{\text{corr}} +$ \Call{Eval}{$q, \text{agent} \oplus \mathcal{D}_{\text{hybrid}}$}
    \State $R^{\text{ret}} \leftarrow R^{\text{ret}} +$ \Call{MRR}{$d_k, q, \mathcal{D}_{\text{hybrid}}$}
\EndFor
\State $R^{\text{corr}} \leftarrow \frac{R^{\text{corr}}}{|\mathcal{Q}_k^{\text{eval}}|}$
\State $R^{\text{ret}} \leftarrow \frac{R^{\text{ret}}}{|\mathcal{Q}_k^{\text{eval}}|}$
\State \Return $\lambda_{\text{corr}} \cdot R^{\text{corr}} + \lambda_{\text{ret}} \cdot R^{\text{ret}}$
\end{algorithmic}
\end{algorithm}

\subsubsection{Integrator Agent Prompt}

We specify the \texttt{IntegAgent} prompt below:

\vspace{2ex}
\begin{lstlisting}[numbers=none]
# Enhanced Documentation Editor Prompt
 
You are a meticulous documentation-level editor specializing in comprehensive technical reference materials. You will be given a list of topic nodes, each containing structured information that must be preserved and enhanced with maximum detail retention.
 
## Input Structure Analysis
Each node contains:
- **Title**: The primary topic identifier
- **Context**: Background information and conceptual foundation
- **How to Use**: Step-by-step instructions, commands, flags, parameters, and implementation details
- **When to Use**: Specific scenarios, conditions, and decision criteria
- **Best Practices**: Expert recommendations, optimization techniques, and common pitfalls to avoid
 
## Detailed Processing Requirements
 
### 1. Information Preservation (Zero Loss Policy)
- **Preserve every technical detail**: All command-line flags, parameter values, configuration options, file paths, URLs, version numbers, and exact syntax
- **Maintain all examples**: Keep every code snippet, sample input/output, file names, directory structures, and command sequences exactly as provided
- **Retain contextual nuances**: Preserve qualifying language like ``typically,'' ``usually,'' ``in most cases,'' ``when available,'' and conditional statements
- **Keep quantitative data**: Preserve all numbers, measurements, timeframes, limits, thresholds, and statistical information
- **Maintain cross-references**: Keep all mentions of related tools, dependencies, prerequisites, and interconnected concepts
 
### 2. Enhanced Detail Extraction
- **Expand abbreviations**: When encountering shortened forms, expand them naturally while preserving the original
- **Surface implicit knowledge**: Make obvious assumptions explicit (e.g., "this requires root permissions," "assumes default configuration")
- **Clarify relationships**: Explicitly describe how different components, options, or steps relate to each other
- **Highlight edge cases**: Emphasize special conditions, exceptions, or unusual scenarios mentioned in the source
- **Elaborate on consequences**: When the source mentions outcomes, expand on both success and failure scenarios
 
### 3. Prose Transformation Guidelines
- **Bullet integration**: Transform each bullet point into 1-3 complete sentences that naturally flow together
- **Technical precision**: Use precise technical vocabulary while maintaining readability
- **Logical flow**: Organize information within each section to follow a logical sequence (setup -> execution -> verification)
- **Contextual embedding**: Weave code snippets and technical terms seamlessly into narrative sentences
- **Comprehensive coverage**: Ensure every sub-bullet, nested item, and parenthetical note becomes part of the prose
 
### 4. Structural Requirements
- **Heading hierarchy**: Use `# Title` for each node's main heading
- **Section order**: Maintain Context -> How to Use -> When to Use -> Best Practices sequence
- **Paragraph organization**: Create substantial paragraphs (3-6 sentences) rather than brief statements
- **Transition quality**: Craft smooth bridges between sections and between different nodes
- **Code formatting**: Preserve all inline code with backticks and maintain proper formatting for code blocks
 
### 5. Quality Assurance Checklist
Before finalizing, verify:
- [ ] Every piece of source information appears in the output
- [ ] All technical specifications, parameters, and examples are intact
- [ ] Code snippets maintain their exact syntax and formatting
- [ ] Prose flows naturally without choppy or fragmented sentences
- [ ] Each section provides comprehensive coverage of its topic area
- [ ] Cross-references and dependencies are clearly explained
- [ ] No section labels or formatting artifacts remain in the prose
 
## Output Specifications
Generate a single, cohesive markdown document that reads as authoritative technical documentation. The result should be comprehensive enough that a reader could successfully implement the described tools or techniques using only the information provided, without referring back to the original nodes.
 
---
 
**Input Nodes:**  
<NODES>
{node_list}
</NODES>
 
---
 
Now, produce the aggregated markdown reference sheet with maximum detail preservation and enhanced clarity.
\end{lstlisting}

\subsection{Hindsight Relabeling: Details}
\label{app:hindsight_details}

Algorithm~\ref{alg:hindsightrelabel} formalizes the hindsight relabeling procedure introduced in Section~\ref{sec:hindsight_method}.

\begin{algorithm}[H]
\caption{HindsightRelabel: Augment trajectory pool via task reframing}
\label{alg:hindsightrelabel}
\begin{algorithmic}[1]
\Require Queries $\mathcal{Q}$, KB $\mathcal{D}$, rubrics
\Ensure Augmented concept-insight pairs $\mathcal{B}_{\text{HR}}$
\State $\mathcal{B}_{\text{HR}} \leftarrow \varnothing$
\State $\mathcal{T}_{\text{pass}}, \mathcal{T}_{\text{fail}} \leftarrow$ \Call{SplitByOutcome}{$\mathcal{Q}, \mathcal{D}$}
\State $\mathcal{T}_{\text{sample}} \leftarrow$ \Call{BalancedSample}{$\mathcal{T}_{\text{pass}}, \mathcal{T}_{\text{fail}}, 0.5$}
\For{each failing trajectory $\tau_f \in \mathcal{T}_{\text{sample}}$}
    \State $\text{summary} \leftarrow$ \Call{Summarize}{$\tau_f$} \Comment{Key decisions, deviation point}
    \State $q' \leftarrow$ \Call{LLM}{$\text{summary}$} \Comment{Generate reframed task}
    \State $\tau' \leftarrow$ \Call{Execute}{$q', \mathcal{D}$} \Comment{Re-run agent on new task}
    \State $\text{label}' \leftarrow$ \Call{Grade}{$\tau'$}
    \State $(c, i) \leftarrow$ \Call{ReflAgent}{$\tau', \text{rubrics}, \text{label}'$}
    \State $\mathcal{B}_{\text{HR}} \leftarrow \mathcal{B}_{\text{HR}} \cup \{(c, i, q')\}$
\EndFor
\State \Return $\mathcal{B}_{\text{HR}}$
\end{algorithmic}
\end{algorithm}

\paragraph{Mechanism.}
We sample an equal number of passing and failing trajectories from the training set. For each failing trajectory, we construct a concise summary capturing the agent's key decisions, intermediate states, and the point of deviation from correct behavior. An LLM then generates a new task description that is \textit{consistent with the agent's observed behavior} rather than the original task. By conditioning task generation on the trajectory, the reframed task remains grounded in realistic execution patterns rather than arbitrary synthetic data.

The agent is re-executed on the generated task. Because the new task was designed to align with what the agent naturally did, re-execution frequently succeeds. Empirically, approximately half of the originally failing trajectories become passing after relabeling, shifting the effective data distribution from a 50/50 pass/fail split to roughly 66\% passing and 33\% failing.

\paragraph{Why relabeled trajectories carry different information.}
The Reflector agent takes as input the trajectory \textit{together with} its correctness label and rubric evaluation. The polarity of this label fundamentally changes the nature of every extracted insight. When the Reflector encounters a failing trajectory, insights are diagnostic: ``this approach does not work.'' When the same behavioral pattern appears in a relabeled trajectory that passes, insights become prescriptive: ``when facing this kind of task, performing X leads to success.'' Without relabeling, competent-but-misaligned behaviors only ever contribute negative signal, and the procedural knowledge they embody is never surfaced as reusable recipes.

A second effect compounds this: re-execution on the reframed task is not guaranteed to reproduce the original trajectory exactly. The agent may follow nearby but distinct action sequences, exploring the local neighborhood of its decision boundary and exposing additional behavioral modes absent from the original dataset.

\paragraph{Effect on KB quality.}
The shift to ${\sim}66/33$ passing/failing has two consequences. First, the KB receives a larger pool of \textit{diverse} successful demonstrations---not redundant with the original passing set, but representing behavioral modes near the agent's decision boundary. Second, the remaining failing trajectories still provide negative signal, preventing brittle overconfidence. The resulting asymmetry biases KB construction toward actionable, prescriptive knowledge while preserving failure coverage.

\subsection{Search Strategy Details}
\label{app:search_strategies}
In addition to EG-MCTS (Section~\ref{mcts}), we evaluate two simpler baselines for KB state search:

\paragraph{Iterative Refinement.}
For each meta-concept $k$, we initialize an empty document $d_k$ and repeatedly invoke the Integrator agent to refine it using newly generated insights. At each step the current document is replaced by the Integrator's output, regardless of quality. This process repeats for up to $B$ iterations (we use $B = E \times K$ to match the total budget of MCTS with expansion width $E$ and depth $K$). Because there is only a single trajectory through document space---no branching, no comparison---any degradation introduced by a single LLM call is permanently committed.
\begin{table*}[]
\centering
\resizebox{\linewidth}{!}{%
\begin{tabular}{lcccc}
\toprule
\textbf{Method} & \textbf{OSWorld Verified} & \textbf{OSWorld Verified} & \textbf{$\tau^2$ Bench} & \textbf{SpreadsheetBench} \\ 
 & GTA1-7B + o3 & GTA1-7B + GPT5.2 & o4-mini & o4-mini\\ \midrule
\method~-Iterative & 53.10 & 62.71 & 57.34 & 42.98 \\
\method~-Iterative---HR & 59.40 & 67.29 & 59.14 & 47.13 \\
\method~-Greedy    & 53.10 & 62.71 & 59.14 & 45.94 \\
\method~-Greedy---HR & 59.40 & 67.29 & 60.71 & 46.66 \\
\method~-MCTS      & 56.80 & 65.83 & 59.14 & 46.80 \\
\method~-MCTS---HR & \textbf{61.70} & \textbf{68.34} & \textbf{62.50} & \textbf{48.50} \\ 
\bottomrule
\end{tabular}%
}
\caption{Ablation over \method~ variants. We compare search strategies (Iterative, Greedy, MCTS) and the effect of Hindsight Relabeling across all three benchmarks.}
\label{tab:ablation}
\end{table*}
\paragraph{Greedy Search.}
At each expansion step for concept $k$, we generate $E$ candidate documents in parallel (matching MCTS width) and select the one with the highest reward. The selected candidate replaces the current document, and the process repeats up to depth $K$. Unlike Iterative Refinement, Greedy Search compares alternatives at each level, providing some robustness to LLM stochasticity. However, it commits entirely to the locally best candidate and never revisits rejected branches, making it susceptible to local optima in complex KB landscapes.

\subsection{Computational Cost and LLM Call Analysis}
\label{app:compute_cost}

We provide a detailed breakdown of the computational resources required by \method~and \method-HR (with Hindsight Relabeling), addressing the cost of KB construction, the MCTS optimization overhead, the additional cost introduced by hindsight relabeling, and the amortized inference-time cost. All experiments were conducted on a single NVIDIA A100 GPU with 4 parallel processes for node expansion.

\subsubsection{Time Complexity}

\method's overall time complexity is:
\[
O\!\bigl(|\mathcal{Q}_{\text{train}}| \cdot T_{\text{LLM}} + |\mathcal{Q}_{\text{train}}| \cdot T_{\text{agent}} + M \cdot |\mathcal{K}| \cdot h \cdot T_{\text{agent}} + M \cdot |\mathcal{K}| \cdot h \cdot T_{\text{eval}}\bigr),
\]
where $T_{\text{LLM}}$ and $T_{\text{agent}}$ denote the wall-clock time for a single LLM call and a full agent episode respectively, $M$ is the number of EG-MCTS iterations, $h$ is the number of candidate expansions per step, $|\mathcal{K}|$ is the number of meta-concepts, and $T_{\text{eval}}$ is the time for one evaluation episode.

In practice, $T_{\text{LLM}} \ll T_{\text{agent}}$ and $T_{\text{eval}} \ll T_{\text{agent}}$, so the perceived complexity reduces to:
\[
O\!\bigl((|\mathcal{Q}_{\text{train}}| + M \cdot |\mathcal{K}| \cdot h) \cdot T_{\text{agent}}\bigr).
\]
Since each \textsc{GenerateInsights} step produces at most $l$ concepts per trajectory, we have $|\mathcal{K}| < l \cdot |\mathcal{Q}_{\text{train}}|$, yielding a final complexity of:
\[
O(M \cdot l \cdot h \cdot |\mathcal{Q}_{\text{train}}| \cdot T_{\text{agent}}).
\]
Empirically, $l \approx 1\text{--}2$ across all benchmarks, confirming that concept growth is sublinear relative to training set size.

\paragraph{Additional cost of Hindsight Relabeling.}
When Hindsight Relabeling is enabled (\method-HR), the initialization phase incurs additional cost from three sources: (i)~summarizing each failing trajectory, (ii)~generating a reframed task description via an LLM call, and (iii)~re-executing the agent on the reframed task. For a balanced sample of $|\mathcal{T}_{\text{fail}}|$ failing trajectories (Algorithm~\ref{alg:hindsightrelabel}), this adds:
\[
O\!\bigl(|\mathcal{T}_{\text{fail}}| \cdot (T_{\text{LLM}} + T_{\text{agent}})\bigr)
\]
to the initialization cost. Since we sample an equal number of passing and failing trajectories, $|\mathcal{T}_{\text{fail}}| \approx \frac{1}{2}|\mathcal{Q}_{\text{train}}|$. The re-execution step dominates ($T_{\text{LLM}} \ll T_{\text{agent}}$), so the effective overhead is approximately $\frac{1}{2}|\mathcal{Q}_{\text{train}}| \cdot T_{\text{agent}}$ additional agent runs during initialization.

Importantly, the MCTS optimization phase itself is not fundamentally more expensive under \method-HR: although the initial insight pool $\mathcal{B}$ is enriched (containing both original and relabeled insights), the number of meta-concepts $|\mathcal{K}|$ after deduplication increases only modestly (see Table~\ref{tab:meta_concept_stats_hr}), and the MCTS search parameters ($M$, $h$, $k$) remain unchanged. The primary cost increase is therefore concentrated in the one-time initialization phase.

\subsubsection{Meta-Concept Statistics}

Table~\ref{tab:meta_concept_stats_hr} reports the number of meta-concepts discovered per benchmark after deduplication, for both \method~and \method-HR. Even for complex domains such as \benchosworld, which spans 10 application categories and 369 tasks, only 23--27 meta-concepts are required. The modest increase under \method-HR reflects the fact that hindsight relabeling surfaces prescriptive knowledge from behavioral modes already present in the training data, which largely map onto existing concept clusters rather than creating entirely new ones.

\begin{table}[h]
\centering
\begin{tabular}{lcccc}
\toprule
\textbf{Benchmark} & \textbf{Training} & \multicolumn{2}{c}{\textbf{Meta-Concepts ($|\mathcal{K}|$)}} & \textbf{Concepts /} \\
 & \textbf{Samples} & \textbf{\method} & \textbf{\method-HR} & \textbf{Trajectory} \\
\midrule
\benchosworld     & 20 & 23 & 27 & 1.15 / 1.35 \\
\benchtau         & 20 &  8 & 10 & 0.40 / 0.50 \\
\benchss          & 30 & 26 & 30 & 0.87 / 1.00 \\
\bottomrule
\end{tabular}
\caption{Number of meta-concepts after deduplication for \method~and \method-HR. Hindsight relabeling introduces a modest increase in concepts (15--25\%) because relabeled trajectories largely consolidate into existing semantic clusters.}
\label{tab:meta_concept_stats_hr}
\end{table}

This consolidation is driven by the semantic clustering in \textsc{DeduplicateConcepts} (Algorithm~\ref{alg:deduplicateconcepts}), which achieves a ratio of approximately 1.15 concepts per training trajectory on \benchosworld~for \method~and 1.35 for \method-HR.

\subsubsection{LLM Call Breakdown}

Table~\ref{tab:llm_calls_hr} provides a per-benchmark breakdown of LLM calls during KB construction for both \method~and \method-HR, alongside agent evaluation costs. The KB construction calls represent a \emph{one-time offline cost} that is amortized across all test episodes.

\begin{table}[h]
\centering
\resizebox{\textwidth}{!}{%
\begin{tabular}{lrrrrrrr}
\toprule
\textbf{Benchmark} & \multicolumn{2}{c}{\textbf{KB Construction LLM Calls}} & \textbf{HR Overhead} & \textbf{Calls / Agent} & \textbf{Test} & \multicolumn{2}{c}{\textbf{Amortized Cost / Episode}} \\
 & \textbf{\method} & \textbf{\method-HR} & & \textbf{Run (Eval)} & \textbf{Episodes} & \textbf{\method} & \textbf{\method-HR} \\
\midrule
\benchosworld     & 1{,}192 & 1{,}680 & +41\% & ${\sim}$21{,}000 & 369   & ${\sim}$3.2 & ${\sim}$4.6 \\
\benchtau         & 1{,}223 & 1{,}715 & +40\% & ${\sim}$22{,}000 & 254   & ${\sim}$4.8 & ${\sim}$6.8 \\
\benchss          & 1{,}397 & 1{,}955 & +40\% & ${\sim}$5{,}400  & 2{,}729 & ${\sim}$0.5 & ${\sim}$0.7 \\
\bottomrule
\end{tabular}%
}
\caption{LLM call statistics for \method~and \method-HR. KB construction calls are a one-time offline cost. Hindsight relabeling adds ${\sim}$40\% more construction calls due to trajectory summarization, task reframing, and agent re-execution. The amortized per-episode overhead remains negligible relative to evaluation-time calls.}
\label{tab:llm_calls_hr}
\end{table}

The ${\sim}$40\% increase in construction calls under \method-HR arises from three sources: (i)~LLM calls for summarizing ${\sim}\frac{1}{2}|\mathcal{Q}_{\text{train}}|$ failing trajectories, (ii)~LLM calls for generating reframed task descriptions, and (iii)~agent re-execution on the reframed tasks, which itself involves multiple LLM calls per episode. The cost increase is proportional to the number of failing trajectories in the balanced sample. Even with this overhead, the amortized per-episode cost remains small: for \benchosworld, \method-HR adds only ${\sim}$4.6 calls per test episode compared to ${\sim}$21{,}000 calls during evaluation, a ratio of 0.02\%.

\paragraph{Why the overhead is justified.}
The key insight behind Hindsight Relabeling is that the same agent behavior receives opposite polarity from the Reflector depending on whether the trajectory is labeled as passing or failing. A failing trajectory produces diagnostic insights (``what went wrong''), while the same behavioral pattern, relabeled as passing under a reframed task, produces prescriptive insights (``what to do''). This dual signal enriches the KB with both cautionary and actionable procedural knowledge. Additionally, re-execution on the relabeled task is not guaranteed to reproduce the original trajectory exactly, exposing nearby behavioral modes that give the Reflector genuinely novel material.

The resulting shift from a 50/50 pass/fail distribution to roughly 66\% passing and 33\% failing biases KB construction toward diverse successful demonstrations, representing latent competences never surfaced as successes in the original data, while retaining enough failure coverage to prevent overconfidence. As shown in Table~\ref{tab:train_acc}, \method-HR achieves the highest train-set accuracy on \benchosworld~(84.00\% vs.\ 78.00\% for \method) and competitive performance on other benchmarks, confirming that the additional construction cost translates into measurably better KB quality.

\subsubsection{MCTS Optimization Overhead}

Table~\ref{tab:mcts_overhead_hr} reports wall-clock times for KB construction under the default MCTS configuration ($M{=}10$, $h{=}3$, $k{=}3$) for both \method~and \method-HR. We additionally compare against greedy construction to isolate the overhead introduced by MCTS exploration and hindsight relabeling.

\begin{table}[h]
\centering
\begin{tabular}{lcccc}
\toprule
\textbf{Benchmark} & \textbf{Greedy} & \textbf{\method~} & \textbf{\method-HR} & \textbf{HR Overhead} \\
 & & \textbf{(MCTS)} & \textbf{(MCTS)} & \textbf{vs.\ \method} \\
\midrule
\benchosworld     & ${\sim}$1h 10min & ${\sim}$1h 25min & ${\sim}$1h 55min & +35\% \\
\benchtau         & ${\sim}$1h 00min & ${\sim}$1h 15min & ${\sim}$1h 45min & +40\% \\
\benchss          & ${\sim}$1h 15min & ${\sim}$1h 30min & ${\sim}$2h 05min & +39\% \\
\bottomrule
\end{tabular}
\caption{Wall-clock time for KB construction (NVIDIA A100, 4 parallel processes). MCTS adds ${\sim}$15--20\% overhead relative to greedy construction. Hindsight relabeling adds a further ${\sim}$35--40\% over \method~due to the additional trajectory summarization, reframing, and re-execution during initialization.}
\label{tab:mcts_overhead_hr}
\end{table}

The MCTS optimization adds approximately 15--20\% wall-clock time compared to greedy construction, while Hindsight Relabeling adds a further ${\sim}$35--40\% on top of \method. The combined overhead of \method-HR over greedy construction ranges from ${\sim}$55--65\%. This overhead is concentrated entirely in the offline initialization phase and is justified by consistent performance gains: for example, \method-HR achieves the strongest results on \benchosworld~(Table~\ref{tab:main_results}), and the additional cost is amortized across hundreds to thousands of test episodes.

\subsubsection{End-to-End Cost Breakdown for \benchosworld}

To provide a concrete picture, we decompose the full pipeline for both \method~and \method-HR on \benchosworld:

\begin{table*}[h]
\centering
\begin{tabular}{p{3cm} p{5.5cm} rr}
\toprule
\textbf{Phase} & \textbf{Computation} & \textbf{\method} & \textbf{\method-HR} \\
\midrule
Trajectory generation         & $20 \times {\sim}2.5$ min/trajectory                   & ${\sim}$50 min  & ${\sim}$50 min  \\
Insight extraction \& dedup   & \textsc{GenerateInsights} + \textsc{DeduplicateConcepts} & ${\sim}$10 min  & ${\sim}$10 min  \\
Hindsight relabeling          & Summarize + reframe + re-execute (${\sim}$10 trajectories) & --           & ${\sim}$30 min  \\
MCTS optimization             & $M \times |\mathcal{K}| \times h$ expansions (4$\times$ parallel)  & ${\sim}$25 min  & ${\sim}$28 min  \\
\midrule
\textbf{Total KB construction} &                                                        & ${\sim}$\textbf{1h 25min} & ${\sim}$\textbf{1h 58min} \\
\midrule
Evaluation   & $369 \times {\sim}2.5$ min/episode                      & \multicolumn{2}{c}{${\sim}$15h 20min} \\
(369 episodes)     &     &  \\
\midrule
\textbf{KB as \% of total}    &                                                        & ${\sim}$1.3\%  & ${\sim}$1.8\% \\
\bottomrule
\end{tabular}
\caption{End-to-end cost breakdown for \benchosworld. The hindsight relabeling phase adds ${\sim}$30 minutes to KB construction, and the slightly larger concept set increases MCTS time marginally. Even with \method-HR, KB construction accounts for less than 2\% of the total compute budget.}
\label{tab:cost_breakdown}
\end{table*}

The hindsight relabeling phase adds approximately 30 minutes: ${\sim}$5 minutes for summarizing 10 failing trajectories and generating reframed tasks (LLM calls), and ${\sim}$25 minutes for re-executing the agent on the 10 reframed tasks. The subsequent MCTS phase is only marginally more expensive for \method-HR (${\sim}$28 min vs.\ ${\sim}$25 min) because the additional meta-concepts (27 vs.\ 23) represent a modest increase in parallel search workload. Overall, \method-HR's KB construction accounts for ${\sim}$1.8\% of the total compute budget, compared to ${\sim}$1.3\% for \method; both negligible relative to the cumulative evaluation cost, while yielding a net reduction of ${\sim}$15\% in execution-time API calls through improved agent efficiency.

\subsubsection{Scalability Considerations}

The number of meta-concepts $|\mathcal{K}|$ is the primary factor governing MCTS cost. In all evaluated benchmarks, $|\mathcal{K}|$ remains below 30 due to effective semantic deduplication, even under \method-HR, which enriches the insight pool. For domains requiring substantially more concepts, several strategies can maintain tractability:

\begin{itemize}[topsep=2pt,itemsep=1pt,parsep=0pt]
    \item \textbf{Increased parallelism:} Since EG-MCTS optimizes each concept independently, scaling to $|\mathcal{K}| = 1{,}000$ requires proportionally more parallel processes but no algorithmic changes.
    \item \textbf{Concept prioritization:} Concepts can be ranked by retrieval frequency or reward variance, and MCTS budget can be allocated preferentially to high-impact concepts.
    \item \textbf{Hierarchical clustering:} For very large concept sets, a two-level hierarchy (concept groups $\to$ individual concepts) can reduce the effective search space.
    \item \textbf{Selective relabeling:} Under \method-HR, the balanced sampling strategy (Algorithm~\ref{alg:hindsightrelabel}) ensures that relabeling cost scales with $\frac{1}{2}|\mathcal{Q}_{\text{train}}|$ rather than with the full training set. For larger training sets, importance sampling can further focus relabeling on the most informative failures.
\end{itemize}
\subsection{\method~Configurations}
\label{app:method_config}

\paragraph{Base LLM Configuration}  
For all \method algorithm steps, we use the OpenAI GPT-4.1-2025-04-14 model as the underlying language model. To balance exploration and stability, we set the temperature to 0.7 for the \texttt{IntegAgent} component to encourage diversity in sampled completions, while all other calls use a temperature of 0.1 for deterministic behavior. The search process employs an expansion width of $e=3$, a maximum search depth of $k=3$, and a maximum of $n=10$ iterations. Reward signals are weighted equally across correctness and retrieval relevance, with $\lambda_{corr} = \lambda_{ret} = 0.5$.

\subsection{Experimental Setup}

\subsubsection{Baseline Methods}
\label{app:baselines}  
We compare \method~against two common reasoning baselines. In-Context Learning augments the input prompt with successful trajectories from related tasks, enabling the model to benefit from relevant prior examples without additional fine-tuning.

\begin{table}[h]
\centering
\vspace{-1em}
\begin{tabular}{lccccc}
\toprule
\textbf{Benchmark} & \textbf{Baseline} & \textbf{Agent-Mem} & \textbf{Cognee} & \textbf{\method} & \textbf{\method-HR} \\ \midrule
OSWorld            & 50.00 & 76.00 & 81.50 & 78.00 & \textbf{84.00} \\
$\tau^2$ Bench     & 50.00 & \textbf{78.33} & 72.00 & \textbf{78.33} & 72.67 \\
SpreadsheetBench   & 50.00 & 63.83 & 71.40 & \textbf{83.39} & 77.49 \\ \bottomrule
\end{tabular}%
\caption{Train-set accuracy across methods.}
\label{tab:train_acc}
\end{table}

\subsubsection{Benchmark Specifications}
\label{app:benchmarks}

\paragraph{\benchosworld: Computer-Use Automation.}
\label{app:osworld}

\textit{Dataset Overview.}
\benchosworld~\citep{osworld} comprises 369 real-world computer-use tasks spanning 10 distinct applications. The benchmark is divided into train and test sets, with the distribution of tasks across domains shown in Table~\ref{tab:osworld-domains}.

\textit{Agent Specifications.}
The UI-Tars-7B variant is a 7B-parameter multimodal transformer fine-tuned for graphical user interface understanding. It operates over an action space of PyAutoGUI commands (e.g., click, type, and key presses). The agent integrates a retrieval module that queries a task-relevant knowledge base using the user-provided description, with the top three retrieved items added to the system prompt. Inputs to the model consist of a screenshot of the active UI paired with the natural language task description.  

The GTA1-7B configuration adopts a two-agent architecture, consisting of a planner and a grounding module. The planner (GTA-1-7B) generates the high-level action sequence, while the grounding module (OpenAI O3) verifies and refines each action before execution. Knowledge retrieval is incorporated differently for each component: the planner performs a single retrieval at the start of execution, which is persisted in its prompt, whereas the grounding module performs dynamic retrievals at each verification step.

\textit{Evaluation Protocol.}
Evaluation uses 134 task-specific scripts designed for automated verification. Success criteria include file state checks (e.g., validating \texttt{.xlsx} or \texttt{.docx} outputs), UI element validation to confirm correct interaction, and process completion checks to ensure that the intended automation sequence was executed successfully.

\begin{table}[h]
\centering
\begin{tabular}{lcc}
\toprule
\textbf{Domain} & \textbf{Test} & \textbf{Train} \\
\midrule
Calc         & 45  & 2 \\
Chrome       & 44  & 2 \\
Writer       & 21  & 2 \\
Gimp         & 24  & 2 \\
Impress      & 45  & 2 \\
Os           & 22  & 2 \\
Thunderbird  & 13  & 2 \\
Multi-apps   & 99  & 2 \\
VLC          & 15  & 2 \\
VSCode       & 21  & 2 \\
\midrule
\textbf{Total} & \textbf{349} & \textbf{20} \\
\bottomrule
\end{tabular}
\caption{Test and Train samples across different domains in \benchosworld.}
\label{tab:osworld-domains}
\end{table}

\paragraph{\benchtau: Interactive Tool Usage.}
\label{app:tau2bench}

\textit{Dataset Overview.}
\benchtau~\citep{barres2025tau2benchevaluatingconversationalagents} extends $\tau$-Bench by introducing bidirectional tool-calling capabilities. The dataset covers multiple service-oriented domains, with domain-level task distributions summarized in Table~\ref{tab:tausq-bench-domains}.

\textit{Domain Characteristics.}
The benchmark spans several domains with distinct task characteristics. The Telecom domain focuses on connectivity troubleshooting, plan modifications, and service activation workflows. The Retail domain includes order processing, return handling, and inventory queries. The Airline domain emphasizes booking modifications and policy-compliant rescheduling scenarios.

\textit{Interaction Settings.}
Two interaction modes are defined. In Easy mode, a human proxy (implemented via GPT-4.1) provides detailed guidance to the agent. The knowledge base is built exclusively from Easy mode trajectories, ensuring high-quality demonstrations for learning. In Hard mode, human intervention is minimized. The knowledge base combines both Easy and Hard trajectories, testing the agent's robustness to underspecified or noisy instructions.

\textit{Evaluation Criteria.}
Task success is measured using domain-specific verification procedures. These include database state checks to validate final outcomes, status checks for confirming service or connection state, natural language verification to ensure correct confirmation statements appear in dialogue, and action matching to confirm that all required steps are completed. Each domain uses a tailored subset of these checks (e.g., Telecom relies primarily on status checks).

\begin{table}[h]
    \centering
    \begin{tabular}{lcccc}
    \toprule
    \textbf{Error Type} & \textbf{Baseline} & \textbf{Cognee} & \textbf{Agent-Mem} & \textbf{\method} \\
    \midrule
    Wrong info       & 26 & \underline{28} & 24 & \textbf{16} \\
    Wrong argument   & 22 & 20 & \underline{24} & \textbf{14} \\
    Wrong decision   & 28 & \underline{30} & 26 & \textbf{18} \\
    Partially resolve & 30 & \underline{32} & 28 & \textbf{24} \\
    \midrule
    Total            & 106 & \underline{110} & 102 & \textbf{72} \\
    \bottomrule
    \end{tabular}
    \caption{Error distribution (\%) on \benchtau~Retail. $\downarrow$ is better. Best in \textbf{bold}, worst \underline{underlined}.}
    \label{tab:tau_errs}
\end{table}

\begin{table}[h]
\centering
\begin{tabular}{lcc}
\toprule
\textbf{Domain} & \textbf{Test} & \textbf{Train} \\
\midrule
Telecom        & 105 & 7 \\
Retail         & 105  & 7 \\
Airline        & 44  & 6 \\
\midrule
\textbf{Total} & \textbf{254} & \textbf{20} \\
\bottomrule
\end{tabular}
\caption{Task-wise breakdown for \benchtau~with assumed 2-shot training samples per domain.}
\label{tab:tausq-bench-domains}
\end{table}

\paragraph{\benchss: Real-World Spreadsheet Manipulation.}
\label{app:spreadsheetbench}

\textit{Dataset Overview.}
\benchss~\citep{spreadsheetbench} consists of 912 instructions collected from four major Excel forums and blogs. Each instruction is paired with spreadsheets reflecting authentic, complex user scenarios, often containing multiple tables and non-standard relational structures. The dataset totals 2,729 test cases, averaging three per instruction. A breakdown of cell-level and sheet-level manipulations is shown in Table~\ref{tab:spreadsheetbench-levels}.

\textit{Task Settings.}
The benchmark defines two dimensions of evaluation:  
\begin{itemize}[topsep=2pt,itemsep=1pt,parsep=0pt]
    \item \textbf{Granularity:} Instructions involve either \emph{cell-level} manipulations (specific ranges such as \texttt{D2:D6}) or \emph{sheet-level} manipulations (entire tables or multi-sheet updates).  
    \item \textbf{Evaluation:} Performance is measured using an Online Judge (OJ)-style protocol. The \emph{soft} setting (IOI-style) awards partial credit when only some test cases are solved, while the \emph{hard} setting (ICPC-style) requires solutions to succeed on all test cases.  
\end{itemize}

\textit{Agent Configuration.}
We evaluate \texttt{o4-mini} using a function-calling agent connected to a single Python execution tool. The agent translates natural language instructions into Python code for spreadsheet manipulation (e.g., modifying cells, applying formulas, restructuring tables). After each tool call, all formulas in the spreadsheet are recalculated to ensure consistency before proceeding to the next step. This setup provides a controlled environment to assess reasoning, code generation, and execution robustness across diverse spreadsheet tasks. 

\begin{table}[h]
\centering
\begin{tabular}{lcc}
\toprule
\textbf{Granularity} & \textbf{Instructions} & \textbf{Test Cases} \\
\midrule
Cell-Level & 329 & 986 \\
Sheet-Level & 583 & 1,743 \\
\midrule
\textbf{Total} & 912 & 2,729 \\
\bottomrule
\end{tabular}
\caption{Cell-level vs. sheet-level distribution in \benchss.}
\label{tab:spreadsheetbench-levels}
\end{table}

\subsubsection{KB Construction and Retrieval Details}
\label{app:kb_details}

\paragraph{Training Data Collection}
\begin{itemize}
    \item \textbf{\benchosworld:} 20 successful trajectories (2 per application domain) and 10 for evals.
    \item \textbf{\benchtau:} 20 trajectories balanced across domains and difficulty settings and 10 for evals.
    \item \textbf{\benchss:} Uniformly sample 30 trajectories for training and 10 for evaluation.
\end{itemize}
All numbers are reported on the remaining train set. 

\paragraph{Retrieval Strategy}
\begin{itemize}
    \item \textbf{Query Formation:}  For each task we take in the seed Natural Language query as the retrieval query.
    \item \textbf{Retrieval Count:} We take top-3 documents for all the retrieval steps.
    \item \textbf{Integration Point:} For \benchss~and \benchosworld~we insert retrievals in the system prompt augmentation. For \benchtau~we perform retrieval after each user interaction.
\end{itemize}

\subsection{Analysis and Ablations}

\subsubsection{Exploration of MCTS Parameters}
\label{app:mcts_params}
We evaluate \benchosworld~on two different MCTS parameter settings.
\begin{itemize}
    \item \textbf{Increased Depth:} We keep maximum width of the tree as 3 and depth as 10 with max number of iterations as 25. We observe that the knowledge base over-optimizes on the train set leading to a poorer performance on test set. 
    \item \textbf{Increased Width:} We reverse the parameters where depth is 3 and maximum width is 10 with max iterations 25. We observe many different styles of KBs are generated storing very similar information; these different styles lead to varied performance on both eval and test set, notifying the importance of state search.
\end{itemize}
We report these numbers in Table~\ref{tab:mcts-table}.

\begin{table}[h]
\resizebox{\textwidth}{!}{%
\begin{tabular}{lcccc}
\hline
        & \textbf{Baseline} & \textbf{max\_width=3, max\_depth=3} & \textbf{max\_width=3, max\_depth=10} & \textbf{max\_width=10, max\_depth=3} \\ \hline
\benchosworld~Verified 
        & 53.30 & 56.80 & 52.69 & \textbf{58.50} \\ 
\benchtau 
        & 56.63 & 59.14 & 55.20 & \textbf{60.80} \\ 
\benchss 
        & 44.30 & 46.80 & 43.20 & \textbf{47.90} \\ \hline
\end{tabular}%
}
\caption{Ablation of MCTS tree parameters across benchmarks. Increasing width improves performance consistently, while deeper trees without sufficient breadth degrade results.}
\label{tab:mcts-table}
\end{table}

\subsubsection{Generalizability of \method}
\label{sec:generalizability}
To assess the generalizability of \textsc{Brew}, we conduct experiments on the \benchtau~dataset by varying the proportion of training data while keeping a fixed held-out evaluation set comprising 50\% of the data. Specifically, we incrementally increase the amount of training data from 0\% to 50\% in steps of 10\% and report performance on the held-out portion.

\begin{figure}[t]
    \centering
    \includegraphics[width=0.6\linewidth]{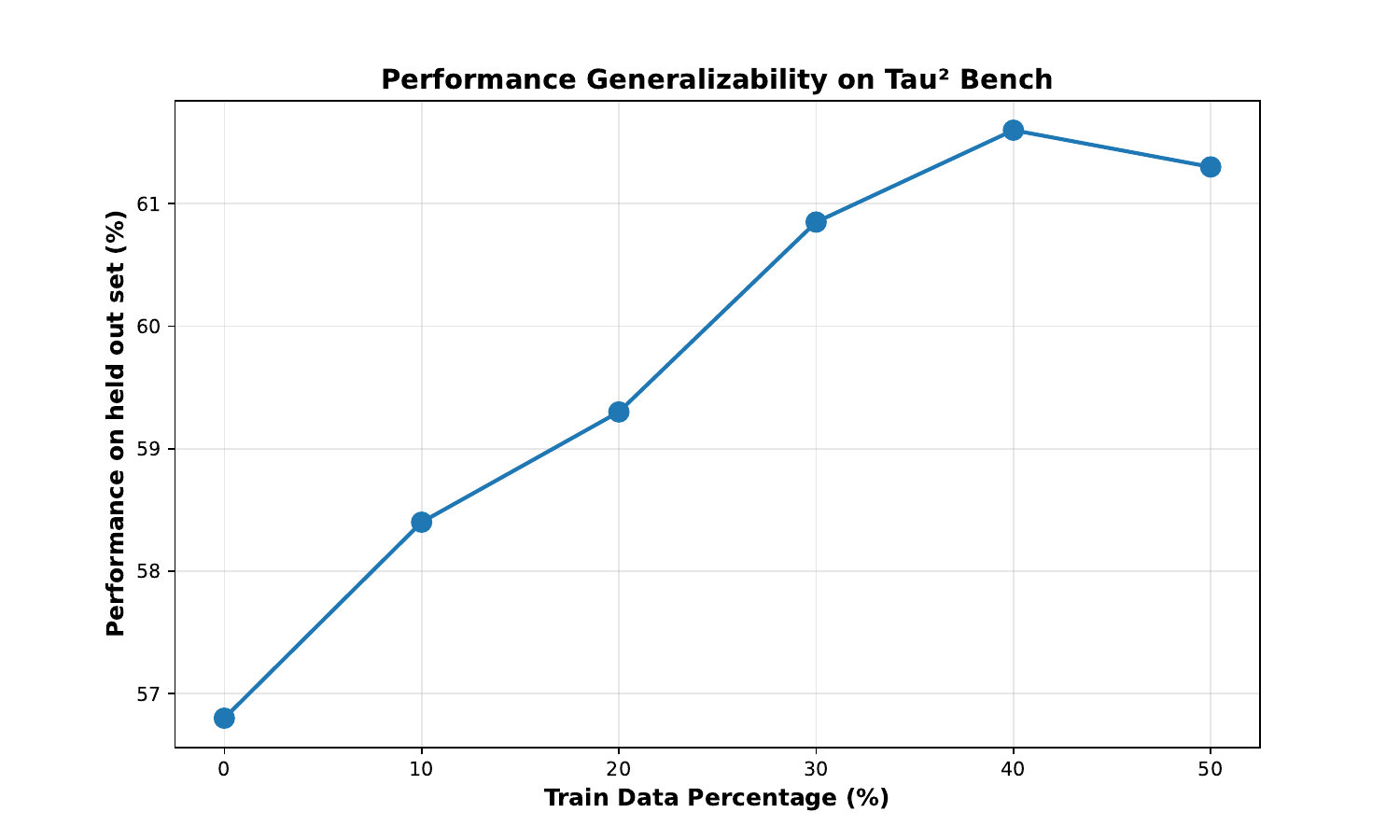}
    \caption{\textbf{Performance generalizability on \benchtau.} The plot shows the relationship between the proportion of training data and performance on the held-out set. \textsc{Brew} exhibits consistent gains with increased training data, demonstrating strong generalization behavior.}
    \label{fig:generalizability}
\end{figure}

As shown in Figure~\ref{fig:generalizability}, model performance improves steadily as more training data is provided, indicating that \textsc{Brew} effectively leverages additional supervision. Notably, the performance gain begins to saturate beyond 40\% of the available training data, suggesting diminishing returns at higher data scales.

\subsubsection{\benchss~Qualitative Analysis}
\label{ssbench_anal}

We identified 109 improvement cases where the \method~KB-enhanced system passed the first test case, the baseline system failed the first test case, and both systems attempted the same task with identical inputs except the KB.
Some common failure patterns identified by \method:

\lstset{
    basicstyle=\ttfamily\small,
    breaklines=true,
    frame=single,
    backgroundcolor=\color{gray!10},
    numbers=none
}
\captionsetup[lstlisting]{labelformat=empty}

\paragraph{Header Detection and Data Structure Preservation.}
Spreadsheet data often lacks explicit headers or contains irregular header structures. Baseline trajectories frequently misinterpret the first data row as column names, corrupting the entire dataset structure.

\subparagraph{Case Study -- Task 280-17.}

\textbf{Task (Simplified):} Remove duplicate rows while keeping last occurrence of each value in column 1.

\textbf{Baseline Approach:} \vspace{0.5em}
\begin{lstlisting}[language=Python]
# Default pandas behavior
df = pd.read_excel(input_path, sheet_name='Sheet1')
last_indices = df.groupby(1, sort=False).apply(lambda grp: grp.index[-1])
filtered_df = df.loc[last_indices.values]
filtered_df.to_excel(output_path, index=False)
\end{lstlisting}

The baseline implementation used default pandas header detection, which promoted the first data row ("DATA 1", 1) to become column headers, fundamentally altering the dataset structure.

\textbf{KB-Enhanced Approach:} \vspace{0.5em}
\begin{lstlisting}[language=Python]
# Explicit header handling guided by KB documentation
df = pd.read_excel(input_path, sheet_name='Sheet1', header=None)
target_col = df.columns[1]
df_filtered = df.drop_duplicates(subset=[target_col], keep='last')
df_filtered.to_excel(writer, sheet_name=name, header=False, index=False)
\end{lstlisting}

The KB system's ``Header Extraction'' documentation guided the conversation toward explicit header validation. The system probed the data structure using \texttt{df.columns.tolist()}, discovered the header issue, and applied appropriate parameters.

\textbf{Impact}: The difference is structural---the baseline output had incorrect data alignment due to header misinterpretation, while the KB output preserved the original data relationships.

\paragraph{Formula Quality and Error Resilience.}
Complex Excel formulas can require domain-specific knowledge about function behavior, particularly regarding blank cell handling and reference types.

\subparagraph{Case Study -- Task 55421.}

\textbf{Task (Simplified):} Create conditional logic for medical appointment scheduling based on multiple criteria including blank date fields.

\textbf{Baseline Formula (Incorrect):} \vspace{0.5em}
\begin{lstlisting}[]
=IF(AND(D2="No Show",E2==""),"CALL PT","")
\end{lstlisting}

\textbf{\method~KB-enhanced Formula (Correct):} \vspace{0.5em}
\begin{lstlisting}[]
=IF(AND($D2="No Show",ISBLANK($E2)),"CALL PT","")
\end{lstlisting}

The baseline approach uses string comparison (\texttt{E2==""}) which fails to properly detect blank cells in Excel. The KB system's validation guidance led to using the \texttt{ISBLANK()} function and absolute references (\texttt{\$D2}, \texttt{\$E2}), resulting in robust formulas that handle edge cases correctly.

\textbf{Impact:} This pattern appeared across multiple formula-intensive tasks, with KB guidance consistently leading to more resilient Excel function usage.

\paragraph{Precise Range Targeting and Scope Compliance.}
Spreadsheet tasks often specify exact cell ranges for modifications. Exceeding these boundaries violates task constraints and can corrupt unrelated data.

\subparagraph{Case Study -- Task 12864.}

\textbf{Task (Simplified):} Use INDEX/MATCH formulas to populate Sheet2!B2:B12 with lookups from Sheet1.

\textbf{Scope Violation:}

The baseline system wrote to both columns B and C:
\begin{lstlisting}[language=Python, caption=Baseline Implementation (Scope Violation)]
# Writes to both B and C columns, violating answer_position
row[1].value = deal_lookup.get(existing_field, "")  # Column B
row[2].value = type_lookup.get(existing_field, "")  # Column C
\end{lstlisting}

The KB system strictly adhered to the specified range:
\begin{lstlisting}[language=Python, caption=KB Implementation (Compliant)]
# Writes formulas only to B2:B12 as specified
formula = f"=INDEX(Sheet1!$B:$B, MATCH($A{row}, Sheet1!$D:$D, 0))"
ws2.cell(row=row, column=2).value = formula  # Only column B
\end{lstlisting}

\textbf{Additional Differences}:
\begin{itemize}
    \item \textbf{Implementation approach}: Baseline used Python dictionaries; KB used actual Excel INDEX/MATCH formulas as requested.
    \item \textbf{Range verification}: KB included diff checking to confirm only target cells changed.
    \item \textbf{Documentation reference}: KB conversation explicitly mentioned ``Column Selection'' guidance.
\end{itemize}

\paragraph{Systematic Validation and Verification.}
Spreadsheet manipulation errors often cascade, making post-execution validation critical for reliability. KB-enhanced conversations consistently included verification steps:
\begin{itemize}
    \item Diff generation to confirm only target cells changed
    \item Output file reloading and inspection
    \item Formula validation through sample execution
    \item Explicit reference to ``Difference in State'' documentation
\end{itemize}

This validation mindset prevented many edge case failures that appeared in baseline implementations.

\paragraph{Discussion.}

\begin{figure}
    \centering
    \includegraphics[width=1\linewidth]{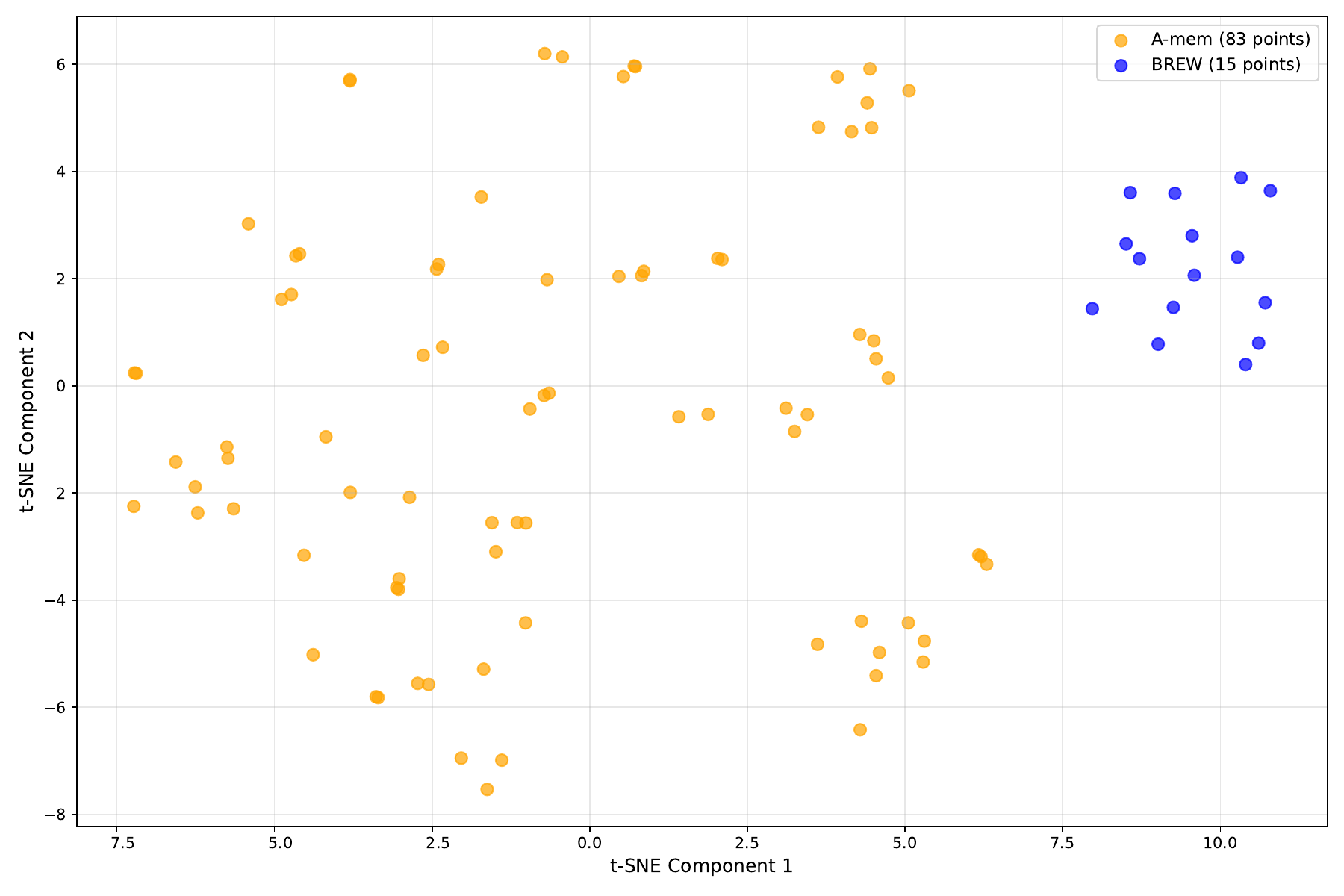}
    \caption{t-SNE plot of knowledge learned by \method~(blue), an experientially learning algorithm, and A-mem (yellow), an agentic tool-based memory storage that relies on the LLM to take memory save actions, on \benchss.}
    \label{fig:ssbench_tsne}
\end{figure}

Automated spreadsheet manipulation presents unique challenges due to the diversity of data structures, formatting conventions, and domain-specific requirements. While large language models demonstrate general competency in coding tasks, they often struggle with the nuanced requirements of spreadsheet operations such as header detection, precise range targeting, and formula construction. Our analysis demonstrates that knowledge base impact extends beyond general process improvement. The documented cases show specific technical guidance that directly prevents implementation errors. For instance, the distinction between \texttt{E2==""} and \texttt{ISBLANK(E2)} represents domain-specific Excel knowledge that a general-purpose system might lack. 

Figure~\ref{fig:ssbench_tsne} shows the t-SNE clustering of knowledge learned by A-mem and \method. The \method~KB is highly structured and domain-adaptive (see Appendix~\ref{prompt:ssbench_kb} for the learned KB). The embedding was done using \texttt{text-embedding-3-large}, where every A-mem entry and every keyword heading in \method~represents a point in the t-SNE embedding space.

The t-SNE visualization of the KB concept embeddings reveals that the knowledge learned by the \method~framework is semantically coherent, forming a dense cluster. This is in stark contrast to the sparse and diffuse memory representation of the A-mem baseline, suggesting that \method's distillation process produces a more structured and less noisy knowledge base.
Analyzing the KB content further reveals that the structure of this knowledge is highly adaptive to the agent's environment. For \benchss, the KB is dominated by abstract formula logic and declarative facts. For \benchosworld, it is primarily procedural, containing step-by-step instructions for GUI navigation. For \benchtau, the KB contains a blend of high-level strategy and procedural tool-use instructions, reflecting the dual demands of conversational reasoning and API execution. This demonstrates that the agent is not simply memorizing trajectories but is building a structured, internal model of its environment tailored to the specific challenges it faces.

\subsubsection{Qualitative Analysis of \method-Generated Knowledge Bases}

This section presents a comprehensive qualitative analysis of knowledge bases generated through the \method~technique applied to two distinct agent training environments: \benchosworld~and \benchtau. The analysis examines knowledge representation patterns, procedural sophistication, and domain-specific learning characteristics extracted from CUA agent behaviors, providing insights into the effectiveness and scope of knowledge distillation techniques across diverse task environments.

\paragraph{Base Structure and Organization.}

\textit{Schema Consistency and Evolution.}
Both knowledge bases demonstrate consistent structural schemas, though adapted to their respective domains. The \benchosworld~KB employs a four-part schema (contextual triggers, procedural steps, extended capabilities, concrete instantiation), while the \benchtau~KB extends this to a five-part structure, adding explicit purpose rationale (``Why to use it''). This evolution suggests that BREW adapts its extraction patterns to domain-specific requirements---conversational commerce demands explicit justification for actions due to customer interaction contexts.

\textit{Taxonomic Organization Principles.}
The \benchosworld~KB reveals a \textbf{capability-based taxonomy} organized around computational tasks: file operations, document processing, inter-application workflows, and data visualization. Each category represents a distinct computational domain with specific tool requirements and interaction patterns. In contrast, the \benchtau~KB employs a \textbf{lifecycle-based taxonomy} structured around transactional states: order creation, modification, fulfillment, and post-delivery operations. This organizational difference reflects fundamental domain characteristics---desktop automation focuses on tool orchestration, while conversational commerce centers on process management.

\textit{Hierarchical Task Decomposition.}
Both KBs demonstrate sophisticated hierarchical reasoning, but through different decomposition strategies. \benchosworld~exhibits \textbf{technical decomposition}, breaking complex operations like ``Create Charts from Data'' into constituent technical steps (data selection, chart insertion, customization, formatting). \benchtau~shows \textbf{process decomposition}, structuring operations like order modification into authentication, validation, confirmation, and execution phases. This suggests BREW successfully identifies domain-appropriate decomposition strategies rather than applying uniform patterns.

\textit{Knowledge Boundary Definition.}
Both KBs explicitly encode operational boundaries, but through contrasting mechanisms. \benchosworld~boundaries are \textbf{capability-constrained}---determined by available applications and system resources. \benchtau~boundaries are \textbf{policy-constrained}---explicitly defined through ``Deny Unsupported Request'' patterns and escalation protocols. This difference highlights how knowledge extraction adapts to domain-specific constraint types.

\paragraph{Procedural Knowledge Grounding.}

\textit{Context-Dependent Action Selection.}
Both domains demonstrate sophisticated context awareness, but grounded in different environmental factors. \benchosworld~exhibits \textbf{application-context sensitivity}, where identical operations (e.g., image insertion) require different procedures across LibreOffice Writer, Impress, GIMP, and Thunderbird. The agent learned application-specific affordances and interaction patterns rather than generic command sequences. \benchtau~demonstrates \textbf{state-context sensitivity}, where available actions depend on order status (pending vs.\ delivered), payment methods, and authentication levels. This reveals learned understanding of business process constraints and temporal operation windows.

\textit{Error Prevention and Validation Workflows.}
Both KBs incorporate sophisticated error prevention mechanisms, but grounded in domain-specific failure modes. \benchosworld~emphasizes \textbf{technical validation}: file integrity checks (``confirm the exported file opens correctly''), application state verification, and multi-step confirmation for irreversible operations. \benchtau~emphasizes \textbf{transactional validation}: authentication cascades, confirmation dialogues with standardized templates, and explicit user consent protocols. The emergence of defensive programming practices across both domains suggests these represent fundamental principles of reliable agent behavior.

\textit{State-Dependent Decision Logic.}
The procedural knowledge in both domains demonstrates sophisticated state machine reasoning. \benchosworld~exhibits \textbf{application state awareness}---understanding when applications are ready for input, when files are loaded, and when operations can be safely executed. Window management and application switching reveal learned understanding of desktop metaphors and resource constraints. \benchtau~demonstrates \textbf{business process state awareness}---finite state machine reasoning where order lifecycle states determine available operations. The agent learned that pending orders enable modification while delivered orders unlock return workflows, indicating internalized understanding of business logic constraints.

\textit{Security and Authentication Grounding.}
While \benchosworld~operates in a trusted desktop environment with minimal explicit security concerns, \benchtau~reveals pervasive \textbf{authentication-first paradigms}. Nearly every transactional operation begins with identity verification through email, name, and zip code combinations. The KB demonstrates \textbf{graduated security reasoning}: information retrieval requires basic authentication while financial transactions trigger rigorous verification protocols. This contrast highlights how procedural knowledge adapts to domain-specific security requirements.

\textit{Cross-Application vs.\ Cross-Process Orchestration.}
\benchosworld~demonstrates \textbf{technical orchestration}---coordinating multiple applications (Chrome, LibreOffice suite, File Manager, GIMP) to accomplish complex workflows. The ``Navigate Between Applications'' section reveals learned behaviors for window management, application switching, and resource coordination. \benchtau~exhibits \textbf{process orchestration}---coordinating authentication, validation, confirmation, and execution phases across different operational contexts. Both forms of orchestration require sophisticated temporal reasoning and constraint management, but applied to different environmental complexity types.

\textit{Failure Mode Internalization.}
Both KBs reveal learned understanding of domain-specific failure modes. \benchosworld~incorporates file validation, application crash recovery suggestions, and verification steps for critical operations. \benchtau~includes explicit escalation protocols (``Transfer to Human Agent''), policy compliance mechanisms, and irreversibility warnings for financial operations. The consistent emergence of failure-aware procedures suggests that agents successfully internalize risk assessment and mitigation strategies during training.

\textit{Domain-Specific Communication Patterns.}
The procedural knowledge reveals distinct communication paradigms appropriate to each domain. \benchosworld~procedures are \textbf{task-oriented} with minimal user interaction---focusing on efficient command execution and verification. \benchtau~procedures are \textbf{dialogue-oriented} with standardized customer interaction templates, confirmation protocols, and expectation management communications. This adaptation demonstrates that BREW extracts not just procedural logic but domain-appropriate interaction modalities.

The cross-domain analysis reveals that BREW successfully extracts procedural knowledge that is both \textbf{structurally consistent} (following learnable organizational patterns) and \textbf{contextually grounded} (adapted to domain-specific constraints, failure modes, and interaction requirements). This dual capability suggests significant potential for knowledge transfer across related domains while maintaining appropriate domain-specific adaptations.

\subsection{Exemplar Knowledge Bases}

\subsubsection{Knowledge Base Learned for \benchosworld}
We showcase a small part of the knowledge base learned through \method. This demonstrates three major parts on which each document is aggregated. These parts discuss when to use a piece of information, why to use the information, and how to use the information/tool. 
\vspace{2ex}
\begin{lstlisting}[numbers=none]
## Search and Open Files

**When to use**: Locating documents, spreadsheets, images, or downloads for editing, conversion, or attachment.

### How to Perform
- Open **File Manager (Nautilus)** from launcher or system dock
- Press `Ctrl + F` or click the search icon
- Enter part of filename, full name, or wildcard (`*.pdf`, `report*`)
- Use right-click -> **Open With** to choose the desired application
- Use the sidebar to navigate to **Downloads**, **Documents**, or custom folders

### Additional Actions
- Right-click -> **Properties** to check modification date or file type
- Sort results by Date, Type, or Name from the top-right dropdown
- Use `F2` to rename files inline

### Example
- Task: "Edit the file titled `sales_report_march.ods`"
  - Search for `sales` in File Manager
  - Confirm `.ods` type and open with LibreOffice Calc

...

## Insert Images

**When to use**: Adding visual elements to documents, presentations, emails, or templates.

### How to Perform
- Navigate to **Insert -> Image -> From File** (in Writer, Impress, Thunderbird)
- Select an image file (`.png`, `.jpg`, `.svg`) from the file dialog
- Use drag handles to resize; right-click -> **Wrap** or **Alignment** for layout

### Additional Actions
- In GIMP: **File -> Open as Layers** to insert image as a new layer
- Use drag-and-drop from file manager into open document windows
- Use **Format -> Image** to apply borders, shadows, or color corrections (in Writer/Impress)

### Example
- Task: "Insert the logo.png image into the title slide"
  - Open `.odp` file in Impress -> Go to Slide 1 -> Insert -> Image -> Select `logo.png`

...

## Export as PDF

**When to use**: Required submission format

### How to Perform
- Go to **File -> Export As PDF**
- Choose output folder (usually **Documents** or **Downloads**)
- Click **Save**, then confirm the exported file opens correctly

### Additional Actions
- In GIMP or Impress: choose **File -> Export As**, then select `.pdf` from format list
- Use **Save As** to preserve both editable and exported versions separately

### Example
- Task: "Export the flyer.xcf as a PDF"
  - Open in GIMP -> File -> Export As -> Rename to `flyer.pdf` -> Click Export
\end{lstlisting}

\subsubsection{\method~Knowledge Base for \benchtau}
BREW enables us to learn relevant information for \benchtau~across the domains in a single knowledge base. This knowledge base is helpful to use relevant actions from the action pool.
\vspace{2ex}
\begin{lstlisting}[numbers=none]
    
### Additional Actions

* Inform the user:  
  - Refunds via gift card = immediate.  
  - Refunds via other methods = 5--7 business days.

### Example

* Task: "Cancel a T-shirt order placed yesterday"
  * Validate: Status is `pending`
  * Reason: "no longer needed"
  * Confirm
  * Execute tool call

# Exchange Delivered Order

**When to use**:  
User wants to swap delivered items for a different variant (e.g., size or color).

**Why to use it**:  
To fix sizing or option errors without needing a new purchase.

### How to Perform
- Authenticate user
- Confirm order status is `delivered`
- Get full list of exchange items
> "Please ensure all items for exchange are listed. This step can't be repeated."
- Ask for refund/payment method
- Confirm:
  > "You're exchanging item X for same product, different option. Proceed?"
- On confirmation:
  ```python
  request_exchange(order_id="45678", item_exchanges=[...], payment_method="paypal")
  ```

### Additional Actions

* Mention: An email will be sent with return instructions
* Validate that the new variant is from the same product

### Example

* Task: "Exchange red shirt for blue in Order #45678"
  * Confirm all exchange items
  * Confirm payment method for difference
  * Execute tool call

### Example

* Task: "Show me my last 2 orders"
  * Authenticate
  * Retrieve and present info

# Deny Unsupported Request

**When to use**:  
User asks for an unsupported action (e.g., cancel processed order, exchange to different product type, help another user).

**Why to use it**:  
To stay compliant with platform policy.

### How to Perform
- Politely reject:
  > "I'm sorry, but I can't process that request. It's outside the allowed scope."

### Example

* Task: "Cancel a processed order"
  * Respond with denial message
# Transfer to Human Agent

**When to use**:  
User needs help outside the assistant's permitted capabilities.

**Why to use it**:  
To ensure user gets the right help from trained staff.

### How to Perform
- Make tool call:
  ```python
  transfer_to_human_agents()
  ```
- Then inform user:
  > "YOU ARE BEING TRANSFERRED TO A HUMAN AGENT. PLEASE HOLD ON."

### Example

* Task: "Delete a task"
  * Deny deletion
  * Transfer to human

\end{lstlisting}

\subsubsection{\method~Knowledge Base for \benchss}
\label{prompt:ssbench_kb}
\begin{lstlisting}[numbers=none]
    Header Extraction
1. Detecting Header Rows
Overview:
To accurately identify header rows, scan the initial region of your dataset. This process is crucial for mapping column information for further processing.
 
Approaches:
- Heuristic Checks:
- Look for rows where all cells are strings (e.g., "Name", "Date", "Region", "Amount").
- Identify rows with distinctive formatting such as bold text or background color.
- Example:
| Name | Date | Region | Amount | |-------|-----------|-----------|--------| | John | 2024-01-01| North | 100 |
- Pattern Recognition:
- Use regex to match typical header patterns, such as column names starting with uppercase letters.
- Score candidate rows based on the likelihood of being headers.
- Multi-Table Sheets:
- Detect gaps, empty rows, or separators indicating a new table.
- Assign a Table ID to each detected table for later reference.
 
Edge Cases:
- Merge multi-row headers (e.g., "Sales" over "2024", "2025" becomes "Sales 2024", "Sales 2025").
- Fill in missing headers by inferring from context.
 
2. Assigning and Validating Headers
Overview:
Once headers are detected, assign them programmatically and ensure they match expected schema and data types.
 
Implementation:
- Column Naming:
- Set names in code, e.g., df.columns = ["Name", "Date", "Region", "Amount"].
- Schema Mapping:
- Map headers to a standardized schema, using external files or user prompts.
- Example:
- Raw header: "Amt"; Mapped header: "Amount"
- Quality Checks:
- Detect duplicate or empty headers ("Date", "Date" becomes "Date_1", "Date_2").
- Validate each column's expected data type.
 
3. Automation and Usability Enhancements
Overview:
Enhance usability and automation to streamline header extraction and user interaction.
 
Features:
- Freeze Panes:
- Automatically freeze header rows in Excel for easier navigation.
- Highlighting:
- Use colored formatting to visually distinguish headers.
- Example:
- Yellow fill for header row.
- Documentation:
- Log extraction logic and confidence scores for each detected header.
- Integration:
- Build header extraction into ETL pipelines and record process metadata.
 
Block Detection
1. Identifying Block Boundaries
Overview:
Block detection segments data into logical units or tables.
 
Methods:
- Boundary Detection:
- Find empty rows, repeated labels, or formatting changes.
- Example:
| Name | Amount | |------|--------| | John | 100 | | | | <-- Empty row indicates new block | Name | Amount | | Alice| 200 |
- Machine Learning:
- Train classifiers to detect block boundaries based on cell patterns.
 
Advanced:
- Detect nested blocks or hierarchies using indentation or merged cells.
- Identify summary blocks with keywords like "Total" or "Summary".
 
2. Processing and Tracking Blocks
Overview:
Once blocks are detected, assign IDs and enable block-level analysis.
 
Actions:
- Block ID:
- Assign unique IDs (e.g., Block_001, Block_002).
- Analysis:
- Perform group-by or aggregation within each block.
- Example:
- Sum "Amount" for Block_001: 100 + 150 = 250
 
3. Additional Block Actions
Overview:
Enable modular analysis and reporting at the block level.
 
Features:
- Summary Rows:
- Add computed totals/averages for each block.
- Export/Save:
- Save blocks as separate files or sheets.
- Example:
- Export Block_001 to "block1.csv"
 
Search for Values or Patterns
1. Search Execution Methods
Overview:
Efficiently locate specific values or patterns in your data.
 
Techniques:
- Manual Tools:
- Use Ctrl + F in Excel for quick lookups.
- Programmatic Search:
- Scan all cells using loops or vectorized code.
- Example:
- Find all instances of "North" in the "Region" column.
- Pattern Matching:
- Support exact, wildcard (*Total*), and regex (\d{4}-\d{2}-\d{2} for dates).
 
2. Recording and Highlighting Results
Overview:
Log and visualize search matches for user review.
 
Actions:
- Logging:
- Record coordinates (e.g., Sheet1, Row 3, Col "Region").
- Highlighting:
- Apply conditional formatting to search hits.
 
3. Advanced Search Scenarios
Overview:
Handle complex or large-scale search requirements.
 
Scenarios:
- Merged Cells:
- Search within merged cells or across multiple sheets.
- Export:
- Export found results for further analysis.
- Example:
- Export all rows containing "John" to "john_results.csv"
 
Writeback Results
1. Output Placement
Overview:
Choose where and how to insert results.
 
Options:
- Target Columns:
- Select existing or blank columns for output.
- Appending:
- Add new columns for flags, counts, or statuses.
- Example:
- Add "Approved_Flag" column next to "Status".
 
2. Writing and Styling Results
Overview:
Automate and style the output for visibility.
 
Methods:
- Formulas/Code:
- Use code (e.g., ws.cell(row, col).value = result) to insert results.
- Styling:
- Bold, borders, or colors for output cells.
- Example:
- Green fill for "Success", red for "Error".
 
3. Audit and Protection
Overview:
Maintain the integrity and traceability of results.
 
Measures:
- Lock Columns:
- Prevent edits to output columns.
- Timestamps/User Info:
- Add audit trail for writebacks.
- Example:
- "2024-06-01, User: admin"
 
Difference in State
1. Sheet Comparison
Overview:
Identify changes between input and output sheets.
 
Process:
- Load Sheets:
- Read both sheets into memory.
- Compare Cells:
- Detect differences by position and value.
 
2. Recording and Reporting Differences
Overview:
Log and report all detected changes.
 
Actions:
- Log Mismatches:
- Record cell coordinates and values.
- Example:
- Cell B3: "North" -> "South"
- Export Diff Report:
- List all detected differences for review.
 
3. Visualization and Automation
Overview:
Make changes visible and automate validation.
 
Features:
- Highlight Changes:
- Color code changed cells.
- Automate Checks:
- Integrate diff comparisons into test scripts.
 
Column Selection
1. Selection Criteria
Overview:
Choose relevant columns for analysis.
 
Methods:
- Labels/Indices:
- Select by name or position.
- Dynamic Rules:
- E.g., all numeric columns.
- Assign Roles:
- Example: "ID", "Date", "Metric"
 
2. Preparation and Validation
Overview:
Prepare columns for consistent use.
 
Actions:
- Rename/Relabel:
- Standardize column names.
- Validate Types:
- Ensure columns are of expected type.
- Example:
- "Date" column as datetime.
 
3. Reusability
Overview:
Save and reuse column selections.
 
Features:
- Presets:
- Save selection profiles.
- Downstream Use:
- Use validated columns in subsequent processes.
 
Filter Rows
1. Filtering Methods
Overview:
Refine your dataset with filters.
 
Techniques:
- Spreadsheet Tools:
- Use built-in filters.
- Code Logic:
- Filter with code (e.g., df[df['Status'] == 'Approved']).
- Multiple Criteria:
- Combine conditions (AND/OR).
- Example:
- Status = "Approved" AND Amount > 100
 
2. Helper Columns and Complex Filters
Overview:
Simplify filtering using helper columns.
 
Actions:
- Helper Columns:
- Compute intermediate flags.
- Document Logic:
- Record filtering rules for audit.
 
3. Post-Filter Actions
Overview:
Visualize and export filtered data.
 
Features:
- Highlighting:
- Grey-out filtered-out rows.
- Export:
- Save the filtered dataset.
 
Merge Tables
1. Key-Based Merging
Overview:
Combine tables using shared keys.
 
Techniques:
- Join Operations:
- Use VLOOKUP, JOIN, or code merges.
- Example:
- Merge "Customer_ID" from two tables.
- Align Data:
- Match on columns like "ID", "Name".
 
2. Stack-Based Merging
Overview:
Append tables when keys aren't needed.
 
Methods:
- Vertical Append:
- Combine rows from similar tables.
- Deduplicate:
- Remove duplicate records.
 
3. Tracking and Audit
Overview:
Track source and unmatched records.
 
Actions:
- Source Column:
- Add "Source" to indicate origin.
- Highlight Unmatched:
- Mark or export mismatched rows.
 
Pivot or Unpivot
1. Pivoting Data
Overview:
Summarize data using pivots.
 
Methods:
- PivotTables:
- Group by row/column dimensions.
- Example:
- Sum "Amount" by "Region".
- Aggregation:
- Choose SUM, AVG, COUNT, etc.
 
2. Unpivoting (Melting) Data
Overview:
Reshape data from wide to long format.
 
Techniques:
- Melt Operations:
- Convert columns into rows.
- Example:
-
| Year | Sales_2019 | Sales_2020 | |------|------------|------------|
->
| Year | Sales_Year | Value |
- Flexible Restructuring:
- Selectively unpivot non-ID columns.
 
3. Post-Pivot Actions
Overview:
Prepare pivoted data for export.
 
Features:
- Flatten Pivot Table:
- Convert back to flat for further analysis.
- Reorder/Rename:
- Clarify pivoted fields.
 
Map with Lookup Tables
1. Mapping Techniques
Overview:
Standardize data using lookups.
 
Methods:
- Functions:
- Use VLOOKUP, merge with dictionaries.
- Code-to-Label:
- Example:
- Code "N" -> Label "North"
 
2. Application and Fallbacks
Overview:
Apply lookups and handle missing values.
 
Actions:
- Apply Mappings:
- Across selected columns.
- Handle Missings:
- Use defaults for missing codes.
 
3. Audit and Display
Overview:
Ensure mapping transparency.
 
Features:
- Cache Mappings:
- Store for repeated use.
- Display Codes/Labels:
- Show both for clarity.
 
Fill Missing Data
1. Choosing Fill Methods
Overview:
Impute missing data appropriately.
 
Techniques:
- Forward/Backward Fill:
- Fill gaps with prior/next value.
- Default Values:
- Use fixed placeholder (e.g., 0, "Unknown").
- Contextual Example:
- Dates: Fill missing month with last known month.
 
2. Application and Auditing
Overview:
Apply fills and flag for review.
 
Actions:
- Targeted Filling:
- Apply to specific columns/rows.
- Flag Filled Cells:
- Highlight for later review.
 
3. Documentation
Overview:
Keep fill logic transparent.
 
Features:
- Record Logic:
- Document assumptions and methods.
- Audit Trail:
- Track all changes.
 
Flag Rows or Cells
1. Defining Flag Rules
Overview:
Establish criteria for flagging.
 
Examples:
- Simple Rule:
- Flag where Amount < 0
- Complex Rule:
- Flag where Status = "Pending" and Amount > 1000
 
2. Applying Flags
Overview:
Insert flags and summarize.
 
Actions:
- Flag Column:
- Add "Flag" column with "Yes"/"No".
- Export Flagged Rows:
- Save for further inspection.
 
3. Advanced Flagging
Overview:
Use multiple criteria and document.
 
Features:
- Multi-Criteria:
- Combine several rules for granular checks.
- Notes:
- Document flagging rationale.
 
Sort Data
1. Setting Sort Criteria
Overview:
Organize data for analysis.
 
Options:
- Sort Columns:
- By value, ascending/descending.
- Multi-Level:
- E.g., sort by "Region", then by "Amount".
 
2. Applying Sorts
Overview:
Implement sorting programmatically or manually.
 
Methods:
- Spreadsheet Tools:
- Built-in sort features.
- Code:
- E.g., df.sort_values(['Region', 'Amount'])
 
3. Post-Sort Actions
Overview:
Finalize sorted data.
 
Actions:
- Renumber Rows:
- Update indices.
- Highlight Extremes:
- Mark top/bottom values.
 
Validate Data
1. Validation Checks
Overview:
Ensure data meets required standards.
 
Checks:
- Type:
- Ensure numeric columns contain numbers.
- Range:
- E.g., "Amount" > 0.
- Pattern:
- Date columns match YYYY-MM-DD.
- Business Rule Example:
- "Start Date" < "End Date"
 
2. Marking and Reporting
Overview:
Visualize and report errors.
 
Actions:
- Highlight Invalids:
- Color-code errors.
- Export Summary:
- Table of error counts and locations.
 
3. Integration in Workflow
Overview:
Make validation a routine part of processing.
 
Features:
- Pre-Processing Step:
- Validate before analysis.
- Automation:
- Integrate into data pipelines.
 
Split Sheets or Data
1. Defining Split Rules
Overview:
Segment data for modular analysis.
 
Methods:
- By Category:
- E.g., split by "Region".
- By Date Range:
- E.g., split by year.
 
2. Exporting Segments
Overview:
Save segments for separate use.
 
Actions:
- Export Files:
- "North_Region.csv", "South_Region.csv"
- Consistent Formatting:
- Ensure identical columns and styling.
 
3. Automation and Documentation
Overview:
Automate splitting and track provenance.
 
Features:
- Automation:
- Use scripts/macros for repeated splits.
- Documentation:
- Record rules and export logs.
\end{lstlisting}